\documentclass{article}

\usepackage{PRIMEarxiv}
\usepackage{natbib}

\usepackage{microtype}
\usepackage{graphicx}
\usepackage{subfigure}
\usepackage{booktabs} 
\usepackage{hyperref}

\usepackage{amsmath}
\usepackage{amssymb}
\usepackage{mathtools}
\usepackage{amsthm}

\usepackage[capitalize,noabbrev]{cleveref}

\usepackage[textsize=tiny]{todonotes}

\usepackage{xspace}
\usepackage[utf8]{inputenc} 
\usepackage[T1]{fontenc}    
\usepackage{hyperref}       
\usepackage{url}            
\usepackage{booktabs}       
\usepackage{tcolorbox}
\usepackage{amsfonts}       
\usepackage{nicefrac}       
\usepackage{microtype}      
\usepackage{xcolor}         

\usepackage{amsmath,amsfonts,bm}

\def\eqref#1{equation~\ref{#1}}

\def\1{\bm{1}}

\DeclareMathAlphabet{\mathsfit}{\encodingdefault}{\sfdefault}{m}{sl}
\SetMathAlphabet{\mathsfit}{bold}{\encodingdefault}{\sfdefault}{bx}{n}

\def\A{{\mathcal{A}}}

\newcommand{\E}{\mathbb{E}}

\newcommand{\btheta}{{\boldsymbol{\theta}}}
\newcommand{\thetafull}{{\btheta_{\mathrm{o}}}}
\newcommand{\thetaunl}{{\btheta_{\mathrm{u}}}}

\newcommand{\Fr}{\mathcal{F_{\text{R}}}}
\newcommand{\Dr}{\mathcal{D_{\text{R}}}}
\newcommand{\Df}{\mathcal{D_{\text{F}}}}
\newcommand{\Da}{\mathcal{D_{\text{A}}}}
\newcommand{\Dt}{\mathcal{D_{\text{T}}}}
\newcommand{\D}{\mathcal{D}}

\usepackage{cleveref}
\usepackage{graphicx}
\usepackage{multirow}

\usepackage{amsmath}
\usepackage{amssymb}
\usepackage{mathtools}
\usepackage{amsthm}
\usepackage{siunitx}
\usepackage{algorithm}
\usepackage{algorithmic}

\usepackage{caption}
\usepackage{subcaption}
\usepackage{float}

\usepackage{wrapfig}

\theoremstyle{plain}
\newtheorem{theorem}{Theorem}[section]

\theoremstyle{definition}
\newtheorem{definition}[theorem]{Definition}

\theoremstyle{remark}

\def\E{\mathbb{E}}

\def\ddefloop#1{\ifx\ddefloop#1\else\ddef{#1}\expandafter\ddefloop\fi}
\def\ddef#1{\expandafter\def\csname bb#1\endcsname{\ensuremath{\mathbb{#1}}}}
\ddefloop ABCDEFGHIJKLMNOPQRSTUVWXYZ\ddefloop
\def\ddef#1{\expandafter\def\csname c#1\endcsname{\ensuremath{\mathcal{#1}}}}
\ddefloop ABCDEFGHIJKLMNOPQRSTUVWXYZ\ddefloop
\def\ddef#1{\expandafter\def\csname v#1\endcsname{\ensuremath{\boldsymbol{#1}}}}
\ddefloop ABCDEFGHIJKLMNOPQRSTUVWXYZabcdefghijklmnopqrstuvwxyz\ddefloop

\def\E{\mathbb{E}}

\def\F{\mathcal{F}}

\newcommand{\amun}{\textsc{AMUN}\xspace}

\def\1{\mathds{1}}

\newif\iffeedback
\feedbacktrue

\iffeedback
\newcommand{\varun}[1]{{\color{red}(Varun: #1)}}
\newcommand{\hari}[1]{{\color{olive}(Hari: #1)}}
\newcommand{\ali}[1]{{\color{green}(Ali: #1)}}
\newcommand{\aliq}[1]{{\color{blue}(Ali Q: #1)}}
\else
\newcommand{\varun}[1]{}
\newcommand{\hari}[1]{}
\newcommand{\ali}[1]{}
\newcommand{\aliq}[1]{}
\fi

{\end{itemize}}

\usepackage{soul}

\title{\amun{}: Adversarial Machine UNlearning}

\author{
  Ali Ebrahimpour-Boroojeny \\
  UIUC \\
  \texttt{ae20@illinois.edu} \\
   \And
  Hari Sundaram \\
  UIUC \\
  \texttt{hs1@illinois.edu}
    \And
  Varun Chandrasekaran \\
  UIUC \\
  \texttt{varunc@illinois.edu}
}

\begin{document}

\maketitle

\begin{abstract}

Machine unlearning, where users can request the deletion of a forget dataset, is becoming increasingly important because of numerous privacy regulations. Initial works on ``exact'' unlearning (e.g., retraining) incur large computational overheads. However, while computationally inexpensive, ``approximate'' methods have fallen short of reaching the effectiveness of exact unlearning: models produced fail to obtain comparable accuracy and prediction confidence on both the forget and test (i.e., unseen) dataset. Exploiting this observation, we propose a new unlearning method, Adversarial Machine UNlearning (\amun{}), that outperforms prior state-of-the-art (SOTA) methods for image classification. \amun{} lowers the confidence of the model on the forget samples by fine-tuning the model on their corresponding adversarial examples. Adversarial examples naturally belong to the distribution imposed by the model on the input space; fine-tuning the model on the adversarial examples closest to the corresponding forget samples (a) localizes the changes to the decision boundary of the model around each forget sample and (b) avoids drastic changes to the global behavior of the model, thereby preserving the model's accuracy on test samples. Using \amun{} for unlearning a random $10\%$ of CIFAR-10 samples, we observe that even SOTA membership inference attacks cannot do better than random guessing. 

\end{abstract}
\section{Introduction}
\label{sec:intro}

The goal of \textit{machine unlearning} is to remove the influence of a subset of the training dataset for a model that has been trained on that dataset~\citep{vatter2023evolution}. The necessity for these methods has been determined by privacy regulations such as the European Union’s General Data Protection Act and the California Consumer Privacy Act.
Despite early efforts on proposing ``exact'' solutions to this problem~\citep{cao2015towards,bourtoule2021machine}, the community has favored ``approximate'' solutions due to their ability to preserve the original model's accuracy while being more computationally efficient~\citep{chen2023boundary,liu2024model,fan2023salun}.

Given a training set $\D$ and a subset $\Df \subset \D$ of the samples that have to be unlearned from a model trained on $\D$, recent works on unlearning have emphasized the use of evaluation metrics that measure the similarity to the behavior of the models that are retrained from scratch on $\D - \Df$. However, prior unlearning methods do not effectively incorporate this evaluation criterion in the design of their algorithm. In this paper, we first characterize the expected behavior of the retrained-from-scratch models on $\D - \Df$. Using this characterization, we propose Adversarial Machine UNlearning (\amun{}). \amun{} is a method that, when applied to the models trained on $\D$, replicates that (desired) behavior after a few iterations. The success of \amun{} relies on an intriguing observation: fine-tuning a trained model on the adversarial examples of the training data does not lead to a catastrophic forgetting and instead has limited effect on the deterioration of model's test accuracy. 

Upon receiving a request for unlearning a subset $\Df$ of the training set $\D$, \amun{} finds adversarial examples that are {\em as close as possible} to the samples in $\Df$. It then utilizes these adversarial examples (with the wrong labels) during fine-tuning of the model for unlearning the samples in $\Df$. Fine-tuning the model on these adversarial examples, which are naturally mispredicted by the model, decreases the confidence of the predictions on $\Df$. This decreased confidence of model's predictions on $\Df$ is similar to what is observed in the models that are retrained on $\D - \Df$. The distance of these adversarial examples to their corresponding samples in $\Df$ is much smaller than the distance of $\Df$ to other samples in $\D - \Df$; this localizes the effect of fine-tuning to the vicinity of the samples in $\Df$ and prevents significant changes to the decision boundary of the model and hurting the model's overall accuracy (see \S~\ref{sec:motivation}). 

As we will show in \S~\ref{sec:experiments}, \amun{} outperforms prior state-of-the-art (SOTA) unlearning methods~\citep{fan2023salun} in unlearning random subsets of the training data from a trained classification model and closes the gap with the retrained models, even when there is no access to the samples in $\D - \Df$ during the unlearning procedure. The code for reproducing the results can be found in \href{https://github.com/Ali-E/AMUN}{https://github.com/Ali-E/AMUN}.

The main contributions of this work are as follow:
\begin{itemize}
\itemsep0em
    \item We observe that neural networks, when fine-tuned on adversarial examples {\em with their wrong labels}, have limited test accuracy degradation. While prior research in adversarial robustness fine-tune the models on these samples with their labels corrected, we are the first to utilize this form of fine-tuning to get lower prediction confidence scores on the training samples that are present in the proximity of those adversarial examples.
    
    \item We introduce a new unlearning method, \amun{}, for classification models that outperforms prior methods. It does so by replicating the behavior of the retrained models on  the test samples and the forget samples. 
   
    \item By comparing \amun{} to existing unlearning methods using SOTA membership inference attacks (MIAs), we show that it outperforms the other methods in unlearning subsets of training samples of various sizes.

\end{itemize}

\subsection{Related Work}

Early works in machine unlearning focused on exact solutions~\cite{cao2015towards,bourtoule2021machine}; those ideas were adapted to unlearning in other domains such as graph neural networks~\cite{chen2022graph} and recommendation systems~\cite{chen2022recommendation}. The extensive computational cost and utility loss resulted in the design of approximate methods. An example is the work of~\citet{ginart2019making}, who provide a definition of unlearning based on differential privacy. Works that followed sought solutions to satisfy those probabilistic guarantees~\citep{ginart2019making,gupta2021adaptive,neel2021descent,ullah2021machine,sekhari2021remember}. However, the methods that satisfy these guarantees were only applied to simple models, such as $k$-means~\cite{ginart2019making} , linear and logistic regression~\cite{guo2019certified,izzo2021approximate}, convex-optimization problems~\cite{neel2021descent}, or graph neural networks with no non-linearities~\cite{chien2022efficient}. 
Additional research was carried out to design more scalable approximate methods, those 
that can be applied to the models that are used in practice, including large neural networks~\cite{golatkar2020eternal,warnecke2021machine,izzo2021approximate,thudi2022unrolling,chen2023boundary,liu2024model,fan2023salun}. However, these approximate methods do not come with theoretical guarantees; their effectiveness are evaluated using membership inference attacks (MIAs). 
MIAs aim to determine whether a specific data sample was used in the training set of a trained model~\citep{shokri2017membership,yeom2018privacy,song2019privacy,hu2022membership,carlini2022membership,zarifzadeh2024low}, and is a common evaluation metric~\citep{liu2024model,fan2023salun}. 
s
For further discussion on related works see Appendix~\ref{apx:related}.

\section{Preliminaries}
\label{sec:prelim}

We begin by introducing the notation we use. We proceed to define various terms in the paper, and conclude by introducing our method.

\subsection{Notation} 
\label{subsec:notation}

Assume a probability distribution $\mathbb{P}_\mathcal{X}$ on the domain of inputs $\mathcal{X}$ and $m$ classes $\mathcal{Y} = \{1,2,\dots,m\}$. We consider a multi-class classifier $\mathcal{F}:\mathcal{X} \rightarrow \mathcal{Y}$ and its corresponding prediction function $f(x)$ which outputs the probabilities corresponding to each class (e.g., the outputs of the softmax layer in a neural network). The loss function for model $\mathcal{F}$ is denoted $\ell_\mathcal{F}: \mathcal{X} \times \mathcal{Y} \rightarrow \mathbb{R_+}$; it uses the predicted scores from $f(x)$ to compute the loss given the true label $y$ (e.g., cross-entropy loss). 

In the supervised setting we consider here, we are given a dataset $\mathcal{D} = \{(x_i, y_i)\}_{i=\{1, \dots, N\}}$ that contains labeled samples $x_i \sim \mathbb{P}_\mathcal{X}$ with $y_i \in \mathcal{Y}$. The model $\F$ is trained on $\D$ using the loss $\ell_\F$ to minimize the empirical risk $\E_{\mathcal{D}} [\ell_\mathcal{F}(x,y)]$ and a set of parameters $\thetafull \sim {\Theta}_\D$ is derived for $\F$; $\Theta_\D$ is the distribution over the set of all possible parameters $\Theta$ when the training procedure is performed on $\D$ due to the potential randomness in the training procedure (e.g., initialization and using mini-batch training). We also assume access to a test set $\Dt$ with samples from the same distribution $\mathbb{P}_\mathcal{X}$. A function $g(x)$ is $L$-Lipschitz if $\| g(x) - g(x^\prime)\|_2 \leq L \| x-x^\prime \|_2, \forall x,x^\prime \in \mathcal{X}$.

\subsection{Definitions}
\label{sec:defs}

\begin{definition}[Attack Algorithm]
\label{def:attack}
    For a given input/output pair $(x,y) \in \mathcal{X} \times \mathcal{Y}$, a model $\mathcal{F}$, and a positive value $\epsilon$, an untargetted attack algorithm $\A_\mathcal{F}(x, \epsilon) = x+\delta_x$ minimizes $\ell_\F(x+\delta_x,y^\prime \neq y)$ such that $\|\delta_x\|_2 \leq \epsilon$, where $y^\prime \in \mathcal{Y}$.
\end{definition}

\begin{definition}[Machine Unlearning]
\label{def:mu}
Given the trained model $\F$, and a subset $\Df \subset \D$ known as the forget set, the corresponding machine unlearning method is a function $\mathcal{M}_{\D,\Df}: \Theta \rightarrow \Theta$ that gets $\thetafull \sim \Theta_\D$ as input and derives a new set of parameters (aka the unlearned model) $\thetaunl \sim \Theta_\Df$, where $\Theta_\Df$ is the distribution over the set of parameters when $\F$ is trained on $\D - \Df$ rather than $\D$. 
\end{definition}

\subsection{Approximate Unlearning}

Using Definition~\ref{def:mu}, it is clear that the most straight-forward, exact unlearning method would be to retrain model $\F$ from scratch on $\D - \Df$; this does not even use $\thetafull$. However, training deep learning models is very costly, and retraining the models upon receiving each unlearning request would be impractical. Thus, approximate unlearning methods are designed to overcome these computational requirements by starting from $\thetafull$ and modifying the parameters to derive $\thetafull^\prime$ s.t.  $\thetafull^\prime \stackrel{\text{d}}{=} \thetaunl$ (i.e., from the same distribution). 
 
In the rest of the paper, we refer to $\Df$ as the forget or unlearning set interchangeably. Its complement, $\Dr = \D -\Df$ is the remain set. We will use the behavior of the models retrained from scratch on $\Dr$ as the goal of approximate unlearning methods, and will refer to them as $\Fr$ for brevity. 

\subsection{Unlearning Settings}
\label{sec:unlear_setting}

Many of the prior methods on approximate unlearning for classification models require access to $\Dr$. However, in practice, this assumption might be unrealistic. The access to $\Dr$ might be restricted, or might be against privacy regulations. 
Prior works do not make a clear distinction based on this requirement when comparing different approximate methods. Therefore, to make a clear and accurate comparison, we perform our experiments (see \S~\ref{sec:experiments}) in two separate settings: one with access to both $\Dr$ and $\Df$, and the other with access to only $\Df$. We report the results for each setting separately. For comparison with prior methods, we adapt them to both settings whenever possible.

\section{Motivation}
\label{sec:methods}

We start by presenting intuition for our proposed unlearning method in \S~\ref{sec:motivation}, and in \S~\ref{sec:adv-finetune} we proceed with describing our observation about fine-tuning a model on its adversarial examples.

\subsection{A Guiding Observation}
\label{sec:motivation}

Before designing a new unlearning method, we would like to first characterize the changes we expect to see after a successful unlearning. Because the retrained models are the gold standard of unlearning methods, we first assess their behavior on $\Dr$, $\Df$, and $\Dt$. To this end, we evaluate the confidence values of $\Fr$ when predicting labels of $\Dr$, $\Df$, and $\Dt$. Since samples in $\Dt$ are drawn from the same distribution as $\D$, we can conclude that samples in $\Dt$ and $\Df$ are from the same distribution. Therefore, we expect $\Fr$ to have similar accuracy and prediction confidence scores on $\Dt$ (test set) and $\Df$. 

\noindent{\bf Results:} Figure~\ref{fig:retrain_conf} in Appendix~\ref{apx:confidence}, shows the confidence scores (see \S~\ref{apx:confidence} for details) for a ResNet-18~\cite{he2016deep} model that has been retrained on $\D - \Df$, where $\D$ is the training set of CIFAR-10~\cite{alex2009learning} and the size of $\Df$ (randomly chosen from $\D$) is $10\%$ and $50\%$ of the size of $\D$ (the first and second sub-figures, respectively). 

\begin{tcolorbox}
\noindent{\bf Key Observation 1:} {\em The main difference between the predictions on $\Dt$ (unseen samples) and $\Dr$ (observed samples) is that the model's predictions are {\em much more confident} for the samples that it has observed compared to the unseen samples.}
\end{tcolorbox}

This basic observation has either been overlooked by the prior research on approximate machine unlearning or has been treated incorrectly. To make the unlearned models more similar to $\Fr$, prior methods have focused on degrading the model's performance on $\Df$ directly by either (a) some variation of fine-tuning on $\Dr$~\citep{warnecke2021machine,liu2024model}, (b) choosing wrong labels for samples in $\Df$ and fine-tuning the model~\citep{golatkar2020eternal,chen2023boundary,fan2023salun}, or (c) directly maximizing the loss with respect to the samples in $\Df$~\citep{thudi2022unrolling}. 
Using the wrong labels for the samples of $\Df$ or maximizing the loss on them make these methods very unstable and prone to catastrophic forgetting~\citep{zhang2024negative} because these samples belong to the correct distribution of the data and we cannot force a model to perform wrongly on a portion of the dataset while preserving it's test accuracy.
For these methods, it is important to use a small enough learning rate along with early stopping to prevent compromising the model's performance while seeking worse prediction confidence values on the samples in $\Df$. Also, most of these methods require access to the set of remaining samples to use it for preventing a total loss of the model's performance (e.g., by continuing to optimize the model on $\Dr$)~\citep{golatkar2020eternal,liu2024model}.

\subsection{Fine-tuning on Adversarial Examples}
\label{sec:adv-finetune}

\begin{figure*}[t!]
\centering
\includegraphics[width=.98\linewidth]{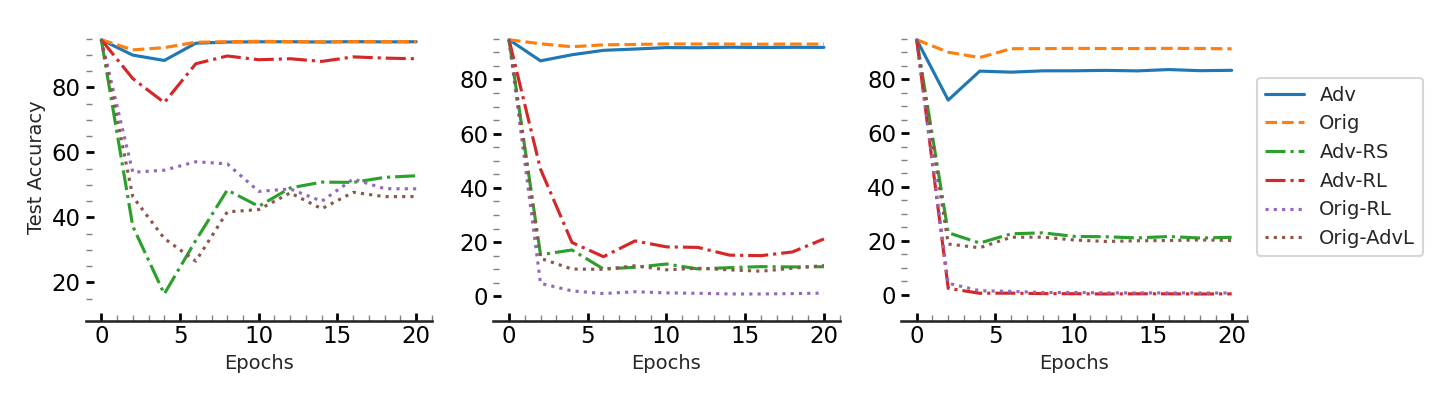}

\caption{{\bf Effect of fine-tuning on adversarial examples.} This figure shows the effect of fine-tuning on test accuracy of a ResNet-18 model that is trained on CIFAR-10, when the dataset for fine-tuning changes (see \S~\ref{sec:ablation} for details). Let $\Df$ contain $10\%$ of the samples in $\D$ and $\Da$ be the set of adversarial examples constructed using Algorithm~\ref{alg:advset}. \texttt{Adv}, from the left sub-figure to right one, shows the results when $\D \cup \Da$, $\Df \cup \Da$, and $\Da$ is used for fine-tuning the model, respectively. \texttt{Orig}, \texttt{Adv-RS}, \texttt{Adv-RL}, \texttt{Orig-RL}, and \texttt{Orig-AdvL} shows the results when $\Da$ for each of these sub-figures is replace by $\Df$, $\Da_{RS}$, $\Da_{RL}$, $\D_{RL}$, and $\D_{AdvL}$, accordingly. As the figure shows, the specific use of adversarial examples with the mis-predicted labels matters in keeping the model's test accuracy because $\Da$, in contrast to the other constructed datasets belong to the natural distribution learned by the trained model.} 
\vspace{-5mm}
\label{fig:fine_tune_10}
\end{figure*}

After training a model $\F$ on $\D$, this model imposes a distribution $f(x)$ (e.g., softmax outputs) for all possible labels $y\in \mathcal{Y}$ given any $x\in \mathcal{X}$. Since the model $\F$ is directly optimized on $\D$, $f(x)$ becomes very skewed toward the correct class for samples in $\D$. For a given sample from $\D$, its adversarial examples (see Definition~\ref{def:attack}) are very close in the input space to the original sample. However, $\F$ makes the wrong prediction on these examples. This wrong prediction is the direct result of the learned parameters $\thetafull$ for the classification model, and these adversarial examples, although predicted incorrectly, belong to the distribution imposed on $\mathcal{X}$ by these learned parameters (i.e., even though that is not the correct distribution, that is what the model has learned).

Now, what happens if we insert one adversarial example $(x_{adv}, y_{adv})$ that corresponds to the sample $(x,y)$ into $\D$ and make an augmented dataset $\D^\prime$ for fine-tuning? Even before fine-tuning starts, the model makes the correct prediction on that (adversarial) example (by predicting the wrong label $y_{adv}$!), but its confidence might not be as high as the samples in $\D$, on which the model has been trained on. Proceeding with
fine-tuning of the model on the augmented dataset increases its confidence on $x_{adv}$ while making the same wrong prediction $y_{adv}$. However, this fine-tuning does not change the model's performance because the newly added sample $(x_{adv}, y_{adv})$ does not contradict the distribution learned by the model. Since $(x,y)\in \D^\prime$, and $x$ and $x_{adv}$ are very close to one another (e.g., very similar images) while having different labels, optimizing the model {\em has to change its decision boundary} in that region of the input space to reach small loss for both of these samples. As a result of this balance, the model tends to decrease its confidence on the original sample compared to the model that was solely trained on $\D$ because there was no opposing components for its optimization on $\D$. Note that $\|x-x_{adv}\| \leq \epsilon$, where $\epsilon$ is often much smaller than the distance of any pairs of samples in $\D$. This helps to localize this change in the decision boundary during fine-tuning, and prevent changes to models' behavior in other regions of the input space~\citep{liang2023towards}. In the following we elaborate on our empirical observations that verify these changes.

\noindent{\bf Setup:} We consider the training set of CIFAR-10 as $\D$ and choose $\Df$ to be a random subset whose size is $10\%$ of $|\D|$. We also compute a set of adversarial examples (using Algorithm~\ref{alg:advset}) corresponding to $\Df$, which we call $\Da$. Fig.~\ref{fig:fine_tune_10} shows the fine-tuning of a trained ResNet-18 model for 20 epochs. In the leftmost sub-figure, the curve presented as \texttt{Orig} represents the test accuracy of the model when it is fine-tuned on $\D$. The curve named \texttt{Adv} is fine-tuned on $\D \cup \Da$, which has a similar test accuracy to \texttt{Orig}.

In the second sub-figure, \texttt{Orig} shows the test accuracy of the model when it is fine-tuned on $\Df$ (two copies of $\Df$ to keep the sample count similar), while \texttt{Adv} represents fine-tuning on $\Df \cup \Da$. As the figure shows, \texttt{Adv} has a small degradation in test accuracy compared to \texttt{Orig}. 

The rightmost sub-figure shows the case where \texttt{Orig} is fine-tuning of the model on $\Df$, and \texttt{Adv} is fine-tuning on only $\Da$. Although the degradation in test accuracy increases for this case, surprisingly we see that the model still remains noticeably accurate despite being fine-tuned on a set of samples that are all mislabeled. See \S~\ref{sec:ablation-finetune} for more details.

\noindent{\bf Results:} As Figure~\ref{fig:fine_tune_10} (and Figure~\ref{fig:fine_tune_50} in Appendix~\ref{apx:ablation-finetune}) shows, the test accuracy of the model does not deteriorate, even when it is being fine-tuned on only $\Da$ (the dataset with wrong labels). See \S~\ref{sec:ablation-finetune} for further details.

\begin{tcolorbox}
\noindent{\bf Key Observation 2:} \textit{Fine-tuning a model on the adversarial examples does not lead to catastrophic forgetting!}
\end{tcolorbox}

\section{Adversarial Machine UNlearning (\amun{})}
\label{sec:amun}

We utilize our novel observation about the effect of fine-tuning on adversarial examples (see \S~\ref{sec:adv-finetune}) to achieve the intuition we had about the retrained models (see \S~\ref{sec:motivation}). We utilize the existing flaws of the trained model in learning the correct distribution, that appear as adversarial examples in the vicinity of the samples in $\Df$, to decrease its confidence on those samples while maintaining the performance of the model. 

Formally, \amun{} uses Algorithm~\ref{alg:advset} to find an adversarial example for any sample in $(x,y) \in \Df$. This algorithm uses a given untargeted adversarial algorithm $\mathcal{A}_\mathcal{F}$, that finds the solution to~\Cref{def:attack}, for finding an adversarial example $x_{adv}$. To make sure $\epsilon$ is as small as possible, Algorithm~\ref{alg:advset} starts with a small $\epsilon$ and runs the attack $\mathcal{A}_\mathcal{F}$; if an adversarial algorithm is not found within that radius, it runs $\mathcal{A}_\mathcal{F}$ with a larger $\epsilon$. It continues to perform $\mathcal{A}_\mathcal{F}$ with incrementally increased $\epsilon$ values until it finds an adversarial example; it then adds it to $\Da$. The algorithm stops once it finds adversarial examples for all the samples in $\Df$. 

The reason behind minimizing the distance of $\epsilon$ for each sample is to localize the changes to the decision boundary of the model as much as possible; this prevents changing the model's behavior on other parts of the input space. For our experiments, we use PGD-50~\citep{madry2017towards} with $\ell_2$ norm bound as $\mathcal{A}_\mathcal{F}$. We set the step size of the gradient ascent in the attack to $0.1\times\epsilon$, which changes with the $\epsilon$ value. More details regarding the implementations of \amun{} and prior unlearning methods and tuning their hyper-parameters can be found in Appendix~\ref{apx:impl_details}. Also, in Appendix~\ref{apx:weak_attack}, we will show how using weaker attacks, such as Fast Gradient Sign Method (FGSM)~\citep{goodfellow2014explaining}, might lead to lower performance of \amun{}.

\begin{algorithm}[H]
\caption{Build Adversarial Set ($\mathcal{F}, \mathcal{A}, \Df, \epsilon_{init}$)}
\label{alg:advset}
\begin{algorithmic}[1]
\STATE {\bfseries Input:} Model $F$, Attack algorithm $A$, Forget set $\Df$, and Initial $\epsilon$ for adversarial attack
\STATE {\bfseries Output:} $\Da$: Adversarial set for $\Df$ 
\STATE $\Da = \{\}$
    \FOR{$(x,y)$ {\bfseries in} $\Df$}
        \STATE $\epsilon = \epsilon_{init}$
        \WHILE{TRUE}
            \STATE $x_{adv} = \mathcal{A}(x, \epsilon)$
            \STATE $y_{adv} = \mathcal{F}(x_{adv})$ 
            \IF{$y_{adv} \, != \, y$}
                \STATE Break
            \ENDIF
            \STATE $\epsilon = 2 \epsilon$
        \ENDWHILE
        \STATE Add $(x_{adv}, y_{adv})$ to $\Da$ 
    \ENDFOR

\STATE {\bfseries Return} $\Da$

\end{algorithmic}

\end{algorithm}

\vspace{-4pt}

Once Algorithm~\ref{alg:advset} constructs $\Da$, \amun{} utilizes that to augment the dataset on which it performs the fine-tuning. If $\Dr$ is available, \amun{} fine-tunes the model on $\Dr \cup \Df \cup \Da$  and when $\Dr$ is not accessible, it performs the fine-tuning on $\Df \cup \Da$. Also, in the setting where the size of the $\Df$ is very large, we noticed some improvement when using only $\Dr \cup \Da$ and $\Da$, for those settings, repsectively.

\subsection{Influencing Factors}

In this section, we derive an upper-bound on the $2$-norm of the difference of the parameters of the unlearned model and the retrained model (which are gold-standard for unlearning) that illuminated the influencing factors in the effectiveness of \amun{}. To prove this theorem, we make assumptions that are common in the certified unlearning literature. The proof is given in~\Cref{apx:proof}.

\begin{theorem}
\label{theorem}

    Let $\mathcal{D} = \{(x_i, y_i)\}_{i=\{1, \dots, N\}}$ be a dataset of $N$ samples and without loss of generality let $(x_n, y_n)$ (henceforth represented as $(x,y)$ for brevity) be the sample that needs to be forgotten and $(x^\prime, y^\prime)$ be its corresponding adversarial example used by \amun{} such that $\|x-x^\prime\|_2 = \delta$. Let $\hat{\mathcal{R}}(w)$ represent the (unnormalized) empirical loss on $\mathcal{D}^\prime = \mathcal{D} \cup \{(x^\prime, y^\prime)\}$ for a model $f$ that is parameterized with $w$. We assume that $f$ is $L$-Lipschitz with respect to the inputs and $\hat{\mathcal{R}}$ is $\beta$-smooth and convex with respect to the parameters. Let $\theta_o$ represent the parameters corresponding to the model originally trained on $\mathcal{D}$ and $\theta_u$ be the parameters derived when the model is trained on $\mathcal{D} - \{(x,y)\}$. We also assume that both the original and retrained models achieve near-$0$ loss on their training sets.  After \amun{} performs fine-tuning on $\mathcal{D}^\prime$ using one step of gradient descent with a learning rate of $\frac{1}{\beta}$ to derive parameters $\theta^\prime$, we get the following upper-bound for the distance of the unlearned model and the model retrained on $\mathcal{D} - \{(x,y)\}$ (gold standard of unlearning):

    \begin{align*}
        \|\theta^\prime - \theta_u\|_2^2 \leq \|\theta_o - \theta_u\|_2^2 + \frac{2}{\beta} (L \delta - C),
    \end{align*}

    \noindent where $C = \ell(f_{\theta_o}(x^\prime), y) + \ell(f_{\theta^\prime}(x^\prime), y^\prime) -\ell(f_{\theta_u}(x), y)  - \ell(f_{\theta_u}(x^\prime), y^\prime) $.

\end{theorem}

According to the bound in Theorem~\ref{theorem}, a lower Lipschitz constant of the model ($L$) and adversarial examples that are closer to the original samples (lower value for $\delta$) lead to a smaller upper bound. A larger value of $C$ also leads to a improved upper-bound. In the following we investigate the factors that lead to a larger value for $C$, which further clarifies some of influencing factors in the effectiveness of \amun{}:

    \begin{itemize}
        \item Higher quality of adversarial example in increasing the loss for the correct label on the original model, which leads to larger value for $\ell(f_{\theta_o}(x^\prime), y)$.
        \item Transferability of the adversarial example generated on the original model to the retrained model to decrease its loss for the wrong label, which leads to a lower value for $\ell(f_{\theta_u}(x^\prime), y^\prime)$.
        \item Early stopping and using appropriate learning rate during fine-tuning phase of unlearning to avoid overfitting to the adversarial example, which does not allow low values for $\ell(f_{\theta^\prime}(x^\prime), y^\prime)$.
        \item The generalization of the retrained model to the unseen samples, which leads to a lower value for $\ell(f_{\theta_u}(x), y) $.

    \end{itemize}

Note that the first two implications rely on the strength of the adversarial example in addition to being close to the original sample. The second bullet, which relies on the transferability of adversarial examples, has been shown to improve as the Lipschitz constant decreases~\citep{ebrahimpourtraining}. The third bullet point is a natural implication which also holds for other unlearning methods that rely on the fine-tuning of the model. The fourth bullet point is not relevant to the unlearning method and instead relies on the fact that the retrained model should have good generalizability to unseen samples; it implies that as the size of $\Df$ increases (i.e., $|\Dr|$ decreases) and the performance of the retrained model decreases, the effectiveness of the unlearning model also decreases. This is also intuitively expected in the unlearning process. Hence, the proved theorem also justifies our earlier intuitions about the need for good quality adversarial examples that are as close as possible to the original samples (which is the goal of Algorithm~\ref{alg:advset}). 

\section{Evaluation Setup}
\label{sec:evaluation}

In this section we elaborate on the details of evaluating different unlearning methods. More details (e.g., choosing the hyper-parameters) can be found in Appendix~\ref{apx:impl_details}.

\subsection{Baseline Methods}
\label{sec:baselines}

We compare \amun{} with \texttt{FT}~\citep{warnecke2021machine}, \texttt{RL}~\citep{golatkar2020eternal}, \texttt{GA}~\citep{thudi2022unrolling}, \texttt{BS}~\citep{chen2023boundary}, $l_1$\texttt{-Sparse}~\citep{liu2024model}, and \texttt{SalUn}~\citep{fan2023salun}. We also combine the weight saliency idea for masking the model parameters to limit the changes to the parameters during fine-tuning with \amun{} and present its results as \amun{}$_{+SalUn}$ (see Appendix~\ref{apx:amun_salun} for more details). We use the same hyper-parameter tuning reported by prior works. For further details, see Appendix~\ref{apx:impl_details}.

\subsection{Evaluation Metrics}
\label{sec:metrics}

The metic used by recent works in unlearning to evaluate the unlearning methods~\citep{liu2024model,fan2023salun} considers the models retrained on $\Dr$ as the goal standard for comparison. They compute the following four values for both the retrained models and the models unlearned using approximate methods:

\begin{itemize}
\itemsep0em
\item \textit{Unlearn Accuracy:} Their accuracy on $\Df$.
\item \textit{Retain Accuracy:} Their accuracy on $\Dr$.
\item \textit{Test Accuracy:} Their accuracy on $\Dt$.
\item \textit{MIA score}: Scores returned by membership inference attacks on $\Df$ 
\end{itemize}

Once these four values are computed, the absolute value of the difference of each of them with the corresponding value for $\mathcal{F}_R$ (the retrained models) is computed. Finally, the average of the four differences (called the {\em Average Gap}) is used as the metric to compare the unlearning methods. 

The MIAs used in the recent unlearning methods by~\citet{liu2024model,fan2023salun} are based on the methods introduced by~\citet{yeom2018privacy,song2019privacy}.
Although these MIAs have been useful for basic comparisons, recent SOTA MIAs significantly outperform their earlier counterparts, albeit with an increase in complexity and computation cost. To perform a comprehensive comparison of the effectiveness of the unlearning methods, we utilized a SOTA MIA called RMIA~\citep{zarifzadeh2024low}, in addition to using the MIAs from prior works. In RMIA, the area under the ROC curve (AUC) of the MIA scores for predicting the training samples from the unseen samples is reported. Recall that in machine unlearning, the samples are split to three sets: $\Dr$, $\Df$, and $\Dt$. For an unlearning method to be effective, as discussed in \S~\ref{sec:motivation}, we expect the AUC of RMIA for distinguishing the samples in $\Df$ from the ones in $\Dt$ to be the same as random guessing ($50\%$ assuming balanced data). As shown in Table~\ref{tab:rmia}, this expectation holds for the models retrained on $\Dr$. 

We report the results of our comparisons for both the MIAs from prior unlearning literature and the new SOTA MIA. We will present the former one as \texttt{MIS}
, and the latter one as \texttt{FT AUC} (the AUC of predicting $\Df$ from $\Dt$).

\subsection{Unlearning Settings} 
\label{sec:settings}

Another important factor missing in the comparisons of the unlearning methods in prior works is the possibility of access to $\Dr$. So, for our experiments we consider two settings, one with access to $\Dr$ and one with access to only $\Df$. We adapt each of the unlearning methods to both of these settings, and perform the comparisons in each of these settings separately. The prior unlearning methods that do not adapt to the setting where there is no access to $\Dr$ \cite{warnecke2021machine,liu2024model} are excluded for the presented results in that setting.

Therefore, we perform different sets of experiments to evaluate the unlearning methods in both settings, and hope this becomes the norm in future works in machine unlearning. In each of these two settings, we evaluate unlearning of $10\%$ or $50\%$ of the samples randomly chosen from $\D$. For all the experiments we train three models on $\D$. For each size of $\D$, we use three random subsets and for each subset, we try three different runs of the unlearning methods. This leads to a total of $27$ runs of each unlearning method using different initial models and subsets of $\D$ to unlearn.

\section{Experiments}
\label{sec:experiments}

We wish to answer the following questions: (1) does \amun{} lead to effective unlearning of any random subset of the samples when evaluated by a SOTA MIA?; (2) does the choice of $\Da$ matter in \amun{}, or can it be replaced with a dataset that contains different labels or different samples that are within the same distance to the corresponding samples in $\Df$?; (3) is \amun{} effective on adversarially robust models?; (4) does the choice of attack method matter in Algorithm~\ref{alg:advset} used by \amun{} and does transferred attack work as well?; and (5) how does \amun{} compare to other unlearning methods when used for performing multiple unlearning requests on the same model?

As a quick summary, our results show that: (1) \amun{} effectively leads to unlearning the samples in $\Df$: after unlearning $10\%$ of the samples of CIFAR-10 from a trained ResNet-18, RMIA cannot do better than random guessing (\S~\ref{sec:results}); (2) If we replace $\Da$ with any of the aforementioned substitutes, the model's accuracy significantly deteriorates, especially when there is no access to $\Dr$ (\S~\ref{sec:ablation-finetune}); (3) \amun{} is as effective for unlearning on models that are adversarially robust (\S~\ref{sec:ablation-robust}); (4) using weaker attack methods, such as FGSM, in \amun{} hurts the effectiveness by not finding the adversarial examples that are very close to the samples in $\Df$. However, they still outperform prior methods (Appendix~\ref{apx:weak_attack}). The transferred adversarial examples are effective as well (Appendix~\ref{apx:transfer_attack}); and (5) \amun{} outperforms other unlearning methods when handling multiple unlearning requests (\S~\ref{sec:adaptive}).

\subsection{Effectiveness of \amun{}}
\label{sec:results}

In this subsection we report the results on the comparisons of \amun{} to other unlearning methods (see \S~\ref{sec:baselines}). We consider the unlearning settings discussed in \S~\ref{sec:settings}, and the evaluation metrics discussed in \S~\ref{sec:metrics}. We use ResNet-18 models trained on CIFAR-10 for this experiment. 

\noindent{\bf Results:} Table~\ref{tab:rmia} shows the results of evaluation using RMIA when the unlearning methods {\em have access to $\Dr$}. Table~\ref{tab:rmia_forgetonly} shows these results when there is {\em no access to $\Dr$}. As the results show, \amun{} {\em clearly outperforms prior unlearning methods in all settings}. This becomes even more clear when there is no access to $\Dr$. Note that, for the models retrained on $\Dr$, the AUC score of RMIA for predicting $\Dr$ from $\Dt$ (which can be considered as the worst case for \texttt{FT AUC} score) are $64.17$ and $69.05$ for unlearning $10\%$ and $50\%$ accordingly. 

We also present the results when MIS is used as the evaluation metric in Tables~\ref{tab:mia} and~\ref{tab:mia_forgetonly} in Appendix~\ref{apx:svc_mia}, which similarly shows \amun{}'s dominance in different unlearning settings. Moreover, we evaluate the combination of \amun{} and \texttt{SalUn} (see Appendix~\ref{apx:amun_salun} for details) and present its results as \amun{}$_{SalUn}$ in these tables. \amun{}$_{SalUn}$ slightly improves the results of \amun{} in the setting where there is no access to $\Dr$, by filtering the parameters that are more relevant to $\Df$ during fine-tuning.

\begin{figure}[t!]
\centering
\includegraphics[width=.98\columnwidth]{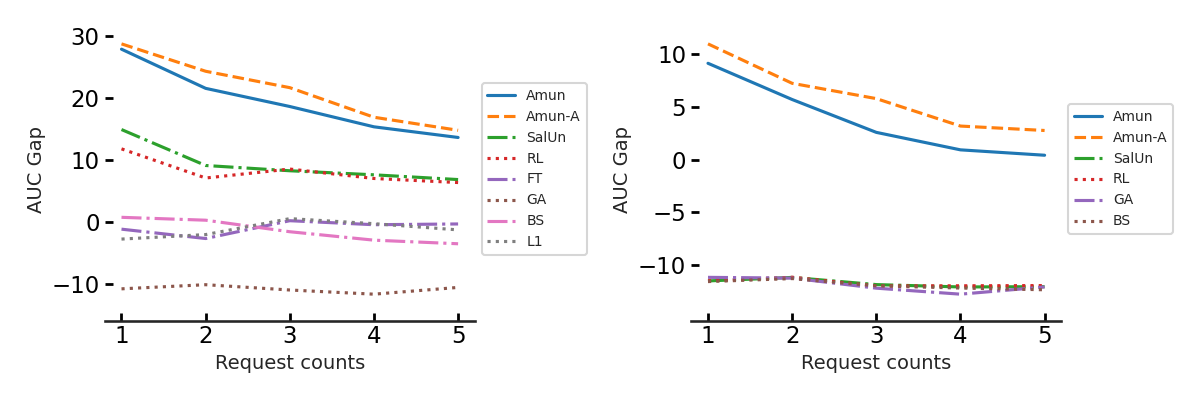}
\caption{{\bf Multiple unlearning requests.} This figure shows the evaluation of unlearning methods when they are used for unlearning for five times and each time on $2\%$ of the training data. We train a ResNet-18 model on CIFAR-10 when $\Dr$ is available (left) and when it is not (right). After each step of the unlearning, we use the MIA scores generated by RMIA to derive the area under the ROC curve (AUC) for $\Dr$ vs. $\Df$ and $\Df$ vs. $\Dt$. The values on the y-axis shows the difference of these two AUC scores. A high value for this gap means the samples in $\Df$ are far more similar to $\Dt$ rather than $\Dr$ and shows a more effective unlearning.} 
\vspace{-5mm}
\label{fig:adaptive}
\end{figure}

\begin{table*}[th!]
\begin{center}
\begin{small}
\begin{sc}
\resizebox{\textwidth}{!}{
\begin{tabular}{@{} l  c c  c  c c | c c c c c @{}}
 \toprule


& \multicolumn{5}{@{}c}{\textbf{Random Forget ($10\%$)}} & \multicolumn{5}{@{}c}{\textbf{Random Forget ($50\%$)}} \\\addlinespace[0.3em]

 \multicolumn{1}{c}{\scriptsize \textbf{}} & 
 \multicolumn{1}{c}{\scriptsize Unlearn Acc} & 
 \multicolumn{1}{c}{\scriptsize Retain Acc} & 
 \multicolumn{1}{c}{\scriptsize Test Acc} & 
 \multicolumn{1}{c}{\scriptsize FT AUC} & 
 \multicolumn{1}{c}{\scriptsize Avg. Gap} & 
 \multicolumn{1}{c}{\scriptsize Unlearn Acc} &
 \multicolumn{1}{c}{\scriptsize Retain Acc} & 
 \multicolumn{1}{c}{\scriptsize Test Acc} & 
 \multicolumn{1}{c}{\scriptsize  FT AUC} & 
 \multicolumn{1}{c}{\scriptsize Avg. Gap} 
 \\\addlinespace[0.3em]

 \cmidrule(r){2-6}
 \cmidrule(r){7-11}



 
 

 Retrain & $94.49$ {\tiny $\pm 0.20$} & $100.0 $ {\tiny $\pm 0.00$} & $94.33$ {\tiny $\pm 0.18$} & $50.00$ {\tiny $\pm 0.42$} & $0.00$ 
 & 
 
 $92.09$ {\tiny $\pm 0.37$}  & $100.0$ {\tiny $\pm 0.00$} & $91.85$ {\tiny $\pm 0.33$} & $50.01$ {\tiny $\pm 0.12$} & $0.00$
 \\\addlinespace[0.3em]
  \cmidrule(r){1-11}

 FT  & $95.16$  { \tiny $\pm 0.29$ }  & $96.64$  { \tiny $\pm 0.25$ }  & $92.21$  { \tiny $\pm 0.27$ }  & $52.08$  { \tiny $\pm 0.34$ }  & $2.06$  { \tiny $\pm 0.10$ } 
 & 
 
 $94.24$  { \tiny $\pm 0.30$ }  & $95.82$  { \tiny $\pm 0.31$ }  & $91.21$  { \tiny $\pm 0.33$ }  & $51.74$  { \tiny $\pm 0.36$ }  & $2.17$  { \tiny $\pm 0.13$ } 
 \\\addlinespace[0.3em]
 
 RL & $95.54$  { \tiny $\pm 0.14$ }  & $97.47$  { \tiny $\pm 0.08$ }  & $92.17$  { \tiny $\pm 0.10$ }  & $51.33$  { \tiny $\pm 0.63$ }  & $1.74$  { \tiny $\pm 0.18$ }  
 & 
 
 $94.83$  { \tiny $\pm 0.44$ }  & $99.79$  { \tiny $\pm 0.04$ }  & $90.08$  { \tiny $\pm 0.16$ }  & $50.78$  { \tiny $\pm 0.14$ }  & $1.38$  { \tiny $\pm 0.09$ } 
 \\\addlinespace[0.3em]

  GA & $98.94$  { \tiny $\pm 1.39$ }  & $99.22$  { \tiny $\pm 1.31$ }  & $93.39$  { \tiny $\pm 1.18$ }  & $60.96$  { \tiny $\pm 2.93$ }  & $4.28$  { \tiny $\pm 0.47$ }  
  & 
 
 $100.00$  { \tiny $\pm 0.00$ }  & $100.00$  { \tiny $\pm 0.00$ }  & $94.65$  { \tiny $\pm 0.07$ }  & $63.39$  { \tiny $\pm 0.26$ }  & $4.62$  { \tiny $\pm 0.05$ } 
 \\\addlinespace[0.3em]

 BS & $99.14$  { \tiny $\pm 0.31$ }  & $99.89$  { \tiny $\pm 0.06$ }  & $93.04$  { \tiny $\pm 0.14$ }  & $57.85$  { \tiny $\pm 1.12$ }  & $3.48$  { \tiny $\pm 0.32$ } 
 & 
  
  $55.24$  { \tiny $\pm 5.11$ }  & $55.67$  { \tiny $\pm 4.90$ }  & $50.16$  { \tiny $\pm 5.28$ }  & $55.19$  { \tiny $\pm 0.42$ }  & $32.01$  { \tiny $\pm 3.86$ } 
 \\\addlinespace[0.3em]

  $l_1$-Sparse & $94.29$  { \tiny $\pm 0.34$ }  & $95.63$  { \tiny $\pm 0.16$ }  & $91.55$  { \tiny $\pm 0.17$ }  & $51.21$  { \tiny $\pm 0.32$ }  & $2.16$  { \tiny $\pm 0.06$ } 
  & 
  
  $98.00$  { \tiny $\pm 0.17$ }  & $98.71$  { \tiny $\pm 0.13$ }  & $92.79$  { \tiny $\pm 0.10$ }  & $54.44$  { \tiny $\pm 0.47$ }  & $2.67$  { \tiny $\pm 0.11$ } 
 \\\addlinespace[0.3em]

  SalUn & $96.25$  { \tiny $\pm 0.21$ }  & $98.14$  { \tiny $\pm 0.16$ }  & $93.06$  { \tiny $\pm 0.18$ }  & $50.88$  { \tiny $\pm 0.54$ }  & $1.44$  { \tiny $\pm 0.12$ } 
  & 

  $96.68$  { \tiny $\pm 0.35$ }  & $99.89$  { \tiny $\pm 0.01$ }  & $91.97$  { \tiny $\pm 0.18$ }  & $50.86$  { \tiny $\pm 0.18$ }  & $1.36$  { \tiny $\pm 0.04$ } 
 \\\addlinespace[0.5em]

 \textbf{Amun}  & $95.45$  { \tiny $\pm 0.19$ }  & $99.57$  { \tiny $\pm 0.00$ }  & $93.45$  { \tiny $\pm 0.22$ }  & $50.18$  { \tiny $\pm 0.36$ }  & $\bf 0.62$  { \tiny $\pm 0.05$ } 
 & 
 
  $93.50$  { \tiny $\pm 0.09$ }  & $99.71$  { \tiny $\pm 0.01$ }  & $92.39$  { \tiny $\pm 0.04$ }  & $49.99$  { \tiny $\pm 0.18$ }  & $\bf 0.33$  { \tiny $\pm 0.03$ } 
 \\\addlinespace[0.3em]

 \textbf{Amun}$_{+SalUn}$ & $95.02$  { \tiny $\pm 0.18$ }  & $99.58$  { \tiny $\pm 0.04$ }  & $93.29$  { \tiny $\pm 0.04$ }  & $50.72$  { \tiny $\pm 0.79$ }  & $\underline{0.68}$  { \tiny $\pm 0.18$ } 
 & 
 
 $93.56$  { \tiny $\pm 0.07$ }  & $99.72$  { \tiny $\pm 0.02$ }  & $92.52$  { \tiny $\pm 0.20$ }  & $49.81$  { \tiny $\pm 0.40$ }  & $\underline{0.36}$  { \tiny $\pm 0.07$ }
 \\\addlinespace[0.3em]

 

 
 
 \bottomrule
\end{tabular}
}
\end{sc}
\end{small}
\end{center}
\caption{\footnotesize {\bf Unlearning with access to $\Dr$.} Comparing different unlearning methods in unlearning $10\%$ and $50\%$ of $\D$. Avg. Gap (see \S~\ref{sec:metrics}) is used for evaluation (lower is better). The lowest value is shown in bold while the second best is specified with underscore. As the results show, \amun{} outperforms all other methods by achieving lowest Avg. Gap and \amun{}$_{SalUn}$ achieves comparable results.\vspace{-1mm}}
\label{tab:rmia}

\end{table*}

\begin{table*}[th!]
\begin{center}
\begin{small}
\begin{sc}
\resizebox{\textwidth}{!}{
\begin{tabular}{@{} l  c c  c  c c | c c c c c @{}}
 \toprule


& \multicolumn{5}{@{}c}{\textbf{Random Forget ($10\%$)}} & \multicolumn{5}{@{}c}{\textbf{Random Forget ($50\%$)}} \\\addlinespace[0.3em]

 \multicolumn{1}{c}{\scriptsize \textbf{}} & 
 \multicolumn{1}{c}{\scriptsize Unlearn Acc} & 
 \multicolumn{1}{c}{\scriptsize Retain Acc} & 
 \multicolumn{1}{c}{\scriptsize Test Acc} & 
 \multicolumn{1}{c}{\scriptsize FT AUC} & 
 \multicolumn{1}{c}{\scriptsize Avg. Gap} & 
 \multicolumn{1}{c}{\scriptsize Unlearn Acc} &
 \multicolumn{1}{c}{\scriptsize Retain Acc} & 
 \multicolumn{1}{c}{\scriptsize Test Acc} & 
 \multicolumn{1}{c}{\scriptsize  FT AUC} & 
 \multicolumn{1}{c}{\scriptsize Avg. Gap} 
 \\\addlinespace[0.3em]

 \cmidrule(r){2-6}
 \cmidrule(r){7-11}



 

 Retrain & $94.49$ {\tiny $\pm 0.20$} & $100.0 $ {\tiny $\pm 0.00$} & $94.33$ {\tiny $\pm 0.18$} & $50.00$ {\tiny $\pm 0.42$} & $0.00$  & 
 
 $92.09$ {\tiny $\pm 0.37$}  & $100.0$ {\tiny $\pm 0.00$} & $91.85$ {\tiny $\pm 0.33$} & $50.01$ {\tiny $\pm 0.12$} & $0.00$
 
 \\\addlinespace[0.3em]
  \cmidrule(r){1-11}

 RL & $100.00$  { \tiny $\pm 0.00$ }  & $100.00$  { \tiny $\pm 0.00$ }  & $94.45$  { \tiny $\pm 0.09$ }  & $61.85$  { \tiny $\pm 0.25$ }  & $4.31$  { \tiny $\pm 0.06$ }  
 & 
 
 $100.00$  { \tiny $\pm 0.00$ }  & $100.00$  { \tiny $\pm 0.00$ }  & $94.57$  { \tiny $\pm 0.14$ }  & $61.99$  { \tiny $\pm 0.10$ }  & $4.29$  { \tiny $\pm 0.03$ }  
 \\\addlinespace[0.3em]

  GA & $4.77$  { \tiny $\pm 3.20$ }  & $5.07$  { \tiny $\pm 3.54$ }  & $5.09$  { \tiny $\pm 3.38$ }  & $49.78$  { \tiny $\pm 0.34$ }  & $68.53$  { \tiny $\pm 2.45$ } 
  &
 
  $100.00$  { \tiny $\pm 0.00$ }  & $100.00$  { \tiny $\pm 0.00$ }  & $92.65$  { \tiny $\pm 0.09$ }  & $63.41$  { \tiny $\pm 0.24$ }  & $5.13$  { \tiny $\pm 0.04$ } 
 \\\addlinespace[0.3em]

 BS & $100.00$  { \tiny $\pm 0.00$ }  & $100.00$  { \tiny $\pm 0.00$ }  & $94.48$  { \tiny $\pm 0.04$ }  & $61.41$  { \tiny $\pm 0.29$ }  & $4.20$  { \tiny $\pm 0.07$ } 
 & 
  
 $100.00$  { \tiny $\pm 0.00$ }  & $100.00$  { \tiny $\pm 0.00$ }  & $94.58$  { \tiny $\pm 0.08$ }  & $62.43$  { \tiny $\pm 0.14$ }  & $4.40$  { \tiny $\pm 0.05$ } 
 \\\addlinespace[0.3em]

  SalUn  & $100.00$  { \tiny $\pm 0.00$ }  & $100.00$  { \tiny $\pm 0.00$ }  & $94.47$  { \tiny $\pm 0.10$ }  & $61.09$  { \tiny $\pm 0.40$ }  & $4.11$  { \tiny $\pm 0.09$ } 
  & 

   $100.00$  { \tiny $\pm 0.00$ }  & $100.00$  { \tiny $\pm 0.00$ }  & $94.59$  { \tiny $\pm 0.12$ }  & $62.45$  { \tiny $\pm 0.37$ }  & $4.40$  { \tiny $\pm 0.07$ } 
 \\\addlinespace[0.5em]

 \textbf{Amun} & $94.28$  { \tiny $\pm 0.37$ }  & $97.47$  { \tiny $\pm 0.10$ }  & $91.67$  { \tiny $\pm 0.04$ }  & $52.24$  { \tiny $\pm 0.23$ }  & $\underline{1.94}$  { \tiny $\pm 0.13$ } 
 & 
 
 $92.77$  { \tiny $\pm 0.52$ }  & $95.66$  { \tiny $\pm 0.25$ }  & $89.43$  { \tiny $\pm 0.19$ }  & $52.60$  { \tiny $\pm 0.22$ }  & $\underline{2.51}$  { \tiny $\pm 0.09$ } 
 \\\addlinespace[0.3em]

 \textbf{Amun}$_{+SalUn}$ & $94.19$  { \tiny $\pm 0.38$ }  & $97.71$  { \tiny $\pm 0.06$ }  & $91.79$  { \tiny $\pm 0.12$ }  & $51.93$  { \tiny $\pm 0.12$ }  & $\bf 1.77$  { \tiny $\pm 0.06$ }
 & 
 
 $91.90$  { \tiny $\pm 0.63$ }  & $96.59$  { \tiny $\pm 0.31$ }  & $89.98$  { \tiny $\pm 0.44$ }  & $52.32$  { \tiny $\pm 0.56$ }  & $\bf 2.00$  { \tiny $\pm 0.17$ }
 \\\addlinespace[0.3em]

 \bottomrule
\end{tabular}
}
\end{sc}
\end{small}
\end{center}
\caption{\footnotesize {\bf Unlearning with access to only $\Df$.}  Comparing different unlearning methods in unlearning $10\%$ and $50\%$ of $\D$. Avg. Gap (see \S~\ref{sec:metrics}) is used for evaluation (lower is better) when only $\Df$ is available during unlearning. As the results show, \amun{}$_{SalUn}$ significantly outperforms all other methods, and \amun{} achieves comparable results. \vspace{-3mm}}
\label{tab:rmia_forgetonly}

\end{table*}

\subsection{Ablation Studies}
\label{sec:ablation}

In this subsection, we first elaborate on the effect of fine-tuning a model on its adversarial examples and compare it to the cases where either the samples or labels of this dataset change (\S~\ref{sec:ablation-finetune}). We then discuss \amun{}'s efficacy on models that are already robust to adversarial examples (\S~\ref{sec:ablation-robust}). We present other ablation studies on using weaker, but faster, adversarial attacks in Algorithm~\ref{alg:advset} (Appendix~\ref{apx:weak_attack}). In Appendix~\ref{apx:transfer_attack}, we utilize transferred adversarial examples for unlearning, as this can expedite handling the unlearning from a newly trained model for which adversarial examples on similar architectures are available.

\subsubsection{Fine-tuning on Adversarial Examples}
\label{sec:ablation-finetune}

We want to verify the importance of $\Da$ (created by Algorithm~\ref{alg:advset}) in preserving the model's test accuracy. To this end, we build multiple other sets to be used instead of $\Da$ when fine-tuning. Let us assume that $\mathcal{A}_\mathcal{F}(x,y) = (x_{adv},y_{adv})$. Then, these other sets are:

\begin{itemize}
\itemsep0em
    \item $\D_{AdvL}$: $\{(x, y_{adv})\}_{\forall (x,y) \in \Df}$
    \item $\D_{RL}$ : $\{(x, y^\prime)$, s.t. $y^\prime \neq y, y_{adv}\}_{\forall (x,y) \in \Df}$
    \item $\Da_{RL}$: $\{(x_{adv}, y^\prime) \text{, s.t.} y^\prime \neq y, y_{adv}\}_{\forall (x,y) \in \Df}$
    \item $\Da_{RS}$: $\{(x^\prime, y_{adv})$, s.t. $x^\prime \sim \mathrm{Uniform(X_\delta)} $, where $X_\delta = \{\forall \hat{x}: \|x_\delta - x\|_2 = \delta\}\}_{\forall (x,y) \in \Df}$
\end{itemize}

In this experiment, we evaluate the effect of fine-tuning on test accuracy of a ResNet-18 model that is trained on CIFAR-10, when $\Da$ is substituted with other datasets that vary in the choice of samples or their labels. We assume that $\Df$ contains $10\%$ of the samples in $\D$ and $\Da$ is the set of corresponding adversarial examples constructed using Algorithm~\ref{alg:advset}. 

\noindent{\bf Results:} In Fig.~\ref{fig:fine_tune_10}, \texttt{Adv}, from the left sub-figure to the right sub-figure, shows the results when $\D \cup \Da$, $\Df \cup \Da$, and $\Da$ is used for fine-tuning the model, respectively. \texttt{Orig}, \texttt{Adv-RS}, \texttt{Adv-RL}, \texttt{Orig-RL}, and \texttt{Orig-AdvL} show the results when $\Da$ for each of these sub-figures is replaced by $\Df$, $\Da_{RS}$, $\Da_{RL}$, $\D_{RL}$, and $\D_{AdvL}$, respectively. As the figure shows, the specific use of adversarial examples with the mispredicted labels matters in keeping the model's test accuracy, especially as we move from the leftmost sub-figure (having access to $\Dr$) to the right one (only using $\Da$ or its substitutes). This is due to the fact that the samples in $\Da$, in contrast to the other constructed datasets, belong to the natural distribution learned by the trained model. Therefore, even if we only fine-tune the ResNet-18 model on $\Da$, we still do not lose much in terms of model's accuracy on $\Dt$. This is a surprising observation, as $\Da$ contains a set of samples with wrong predictions! Fig.~\ref{fig:fine_tune_50} in Appendix~\ref{apx:ablation-finetune} shows similar results when size of $\Df$ is $50\%$ of $|\D|$.

\subsubsection{Adversarially Robust Models}
\label{sec:ablation-robust}

We evaluate the effectiveness of \amun{} when the trained model is adversarially robust. 
One of the most effective methods in designing robust models is adversarial training which targets smoothing the model's prediction function around the training samples~\cite{salman2019provably}. This has been shown to provably enhance the adversarial robustness of the model~\cite{cohen2019certified}. One of the effective adversarial training methods is by using TRADES loss introduced by~\citep{zhang2019theoretically}. We will use adversarially trained ResNet-18 models for unlearning $10\%$ of the samples in CIFAR-10. In addition, we will use another defense mechanism that is less costly and more practical for larger models. There is a separate line of work that try to achieve the same smoothness in model's prediction boundary by controlling the Lipschitz constant of the models~\cite{szegedy2013intriguing}. The method proposed by~\citet{boroojeny2024spectrum} is much faster than adversarial training and their results show a significant improvement in the robust accuracy. We use their clipping method to evaluate the effectiveness of \amun{} for unlearning $10\%$ and $50\%$ of the samples from robust ResNet-18 models trained on CIFAR-10.


\noindent{\bf Results:} Table~\ref{tab:trades_results} in Appendix~\ref{apx:ablation_robust} shows the results for the adversarially trained models. For the models with controlled Lipschitz continuity, the results are shows in Table~\ref{tab:rmia-clipped-remFalse} (no access to $\Dr$) and Table~\ref{tab:rmia-clipped-remTrue} in Appendix~\ref{apx:ablation_robust} (with access to $\Dr$). As the results show, even when there is no access to $\Dr$, \amun{} still results in effective unlearning for adversarially robust models; RMIA does not do better than random guessing for predicting $\Df$ from $\Dt$. As Fig.~\ref{fig:retrain_conf} (right) shows, in the robust models, more than $97\%$ of the adversarial examples are further away from their corresponding training samples, compared to this distance for the original models. However, this does not interfere with the performance of \amun{} because these robust models are smoother and tend to be more regularized. This regularization, which prevents them from overfitting to the training samples is in fact a contributing factor to the improved generalization bounds for these models~\cite{bartlett2017spectrally}. This in itself contributes to enhanced resilience against MIAs. As seen in Tables~\ref{tab:rmia-clipped-remFalse} and~\ref{tab:rmia-clipped-remTrue}, even for the clipped models retrained on $\Dr$, the AUC score of RMIA for predicting $\Dr$ from $\Df$ (\texttt{FR AUC}) is very low, which shows that these smoother models .

\begin{table}[th!]
\begin{center}
\begin{small}
\begin{sc}
\resizebox{0.76\columnwidth}{!}{
\begin{tabular}{@{} l  c c c | c c c @{}}
 \toprule

& \multicolumn{3}{@{}c}{\textbf{Random Forget ($10\%$)}} & \multicolumn{3}{@{}c}{\textbf{Random Forget ($50\%$)}} \\\addlinespace[0.3em]

 \multicolumn{1}{c}{\scriptsize \textbf{}} & 
 \multicolumn{1}{c}{\scriptsize FT AUC} & 
 \multicolumn{1}{c}{\scriptsize FR AUC} & 
 \multicolumn{1}{c}{\scriptsize Test Acc} & 
 \multicolumn{1}{c}{\scriptsize FT AUC} &
 \multicolumn{1}{c}{\scriptsize FR AUC} & 
 \multicolumn{1}{c}{\scriptsize Test Acc} 
 \\\addlinespace[0.3em]

 \cmidrule(r){2-4}
 \cmidrule(r){5-7}

 Retrain & $49.95$ {\tiny $\pm 0.24$} & $54.08 $ {\tiny $\pm 0.16$} & $89.01$ {\tiny $\pm 0.21$} & 
 
 $50.19$ {\tiny $\pm 0.15$}  & $55.61$ {\tiny $\pm 0.05$} & $85.76$ {\tiny $\pm 0.41$}
 
 \\\addlinespace[0.3em]
  \cmidrule(r){1-7}

 \textbf{Amun} & $49.55$ {\tiny $\pm 0.13$} & $54.01$ {\tiny $\pm 0.23$} & $87.55$ {\tiny $\pm 0.44$} & 
 
 $49.64$ {\tiny $\pm 0.31$} & $53.23$ {\tiny $\pm 0.21$} & $87.39$ {\tiny $\pm 0.61$}
 \\\addlinespace[0.3em]
               
 \bottomrule
\end{tabular}
}
\end{sc}
\end{small}
\end{center}
\caption{\footnotesize {\bf Unlearning on adversarially robust models.} Evaluating the effectiveness of \amun{} in unlearning $10\%$ and $50\%$ of the training samples when the models are adversarially robust and there is no access to $\Dr$. For this experiment we use models with controlled Lipschitz constant which makes them provably and empirically more robust to adversarial examples. \vspace{-3mm}}
\label{tab:rmia-clipped-remFalse}

\end{table}

\subsection{Continuous Unlearning}
\label{sec:adaptive}

We evaluate the performance of the unlearning methods when they are used to perform multiple consecutive unlearning from a trained model. This is a desirable capability for unlearning methods because in real-world applications there might be multiple unlearning requests and it is preferred to minimize the number of times that a model needs to be retrained from scratch. 
The setting we envision is as follows: models are updated at each request for unlearning. For \amun{}, this means that $\Da$ is computed on an updated model after each set of unlearning requests (shown as \amun{}\texttt{-A}). 
In addition to comparing \amun{}\texttt{-A} to the other unlearning methods, we also compare it to a version (shown as \amun{}) that computes all the adversarial examples on the original model so it can handle the unlearning requests faster upon receiving them i.e., $\Da$ is not computed on an updated model after each request; the set of requests are batched and $\Da$ is computed on the entire batch. For this experiment, we use a ResNet-18 model trained on training set of CIFAR-10 ($50$K samples). Our goal is to unlearn $10\%$ of the training samples ($5$K), but this time in $5$ consecutive sets of size $2\%$ ($1$K) each. We then evaluate the effectiveness of unlearning at each step using RMIA.

\noindent{\bf Results:} Fig.~\ref{fig:adaptive} shows an overview of the results for both settings of unlearning (with or without access to $\Dr$). This figure shows the effectiveness of unlearning by depicting how the samples in $\Df$ are more similar to the test samples ($\Dt$) rather than the remaining samples ($\Dr$). The value on the y-axis shows the difference of the area under the ROC curve (AUC) for predicting $\Dr$ from $\Df$ and $\Df$ from $\Dt$. For the plots of each of these values separately, see Appendix~\ref{apx:adaptive}. 
\amun{}\texttt{-A} performs better than all the other unlearning methods for all the steps of unlearning. Although \amun{} also outperforms all the prior unlearning methods, it slightly under-performs compared to \amun{}\texttt{-A}. This is expected, as the model's decision boundary slightly changes after each unlearning request and the adversarial examples generated for the original model might not be as effective as those ones generated for the new model. Note that for this experiment, we did not perform hyper-parameter tuning for any of the unlearning methods, and used the same ones derived for unlearning $10\%$ of the dataset presented in \S~\ref{sec:results}. For further discussion of the results see Appendix~\ref{apx:adaptive}.

\section{Conclusions}
\label{sec:conclusions}

\amun{} utilizes our new observation on how fine-tuning the trained models on adversarial examples that correspond to a subset of the training data does not lead to significant deterioration of model's accuracy. Instead, it decreases the prediction confidence values on the the corresponding training samples. By evaluating \amun{} using SOTA MIAs, we show that it outperforms other existing method, especially when unlearning methods do not have access to the remaining samples. It also performs well for handling multiple unlearning requests. This work also raises some questions for future work:  (1) Since SOTA MIA methods fail to detect the unlearned samples, can this method be used to provide privacy guarantees for all the training samples?; (2) Can the same ideas be extended to generative models or Large Language Models?; (3) Can we derive theoretical bounds on the utility loss due to fine-tuning on adversarial examples? 
\clearpage
\section{Acknowledgment}

This work used Delta computing resources at National Center for Supercomputing Applications through allocation CIS240316 from the Advanced Cyberinfrastructure Coordination Ecosystem: Services \& Support (ACCESS) program~\cite{boerner2023access}, which is supported by U.S. National Science Foundation grants \#2138259, \#2138286, \#2138307, \#2137603, and \#2138296.

\bibliographystyle{plainnat}
\bibliography{main}

\appendix

\onecolumn

\section{Proofs}
\label{apx:proof}

Here we provide the proof of Theorem~\ref{theorem}:

\begin{proof}
    As we perform the unlearning by fine-tuning and performing a gradient descent update to $\theta_o$, we have:
        $\theta^\prime =  \theta_o - \frac{1}{\beta} \nabla \hat{\mathcal{R}} (\theta_o)$.
    Therefore, we can write:

    \begin{align*}
        \|\theta^\prime - \theta_u\|_2^2  &= \|\theta_o - \frac{1}{\beta} \nabla \hat{\mathcal{R}} (\theta_o) - \theta_u\|_2^2 \\
        &=  
        \|\theta_o - \theta_u\|_2^2 - \frac{2}{\beta} \langle \nabla \hat{\mathcal{R}} (\theta_o), \theta_o - \theta_u \rangle + \frac{1}{\beta^2} \|\nabla \hat{\mathcal{R}} (\theta_o)\|_2^2 \\
        &\leq 
        \|\theta_o - \theta_u\|_2^2 + \frac{2}{\beta} (\hat{\mathcal{R}} (\theta_u) - \hat{\mathcal{R}} (\theta_o)) + \frac{2}{\beta} (\hat{\mathcal{R}} (\theta_o) - \hat{\mathcal{R}} (\theta^\prime)) \\
        &= 
        \|\theta_o - \theta_u\|_2^2 + \frac{2}{\beta} (\hat{\mathcal{R}} (\theta_u) - \hat{\mathcal{R}} (\theta^\prime)),
    \end{align*}

    \noindent where the inequality is derived by using the smoothness property ($\|\nabla \hat{\mathcal{R}} (\theta_o)\|_2^2 \leq 2\beta(\hat{\mathcal{R}} (\theta_o) - \hat{\mathcal{R}} (\theta^\prime))$) and the convexity assumption which leads to the inequality: $\hat{\mathcal{R}} (\theta_o)) \geq \hat{\mathcal{R}} (\theta_u) + \langle \nabla \hat{\mathcal{R}} (\theta_o), \theta_o - \theta_u \rangle$.

    Next, we derive an upper-bound for $\hat{\mathcal{R}} (\theta_u) - \hat{\mathcal{R}} (\theta^\prime)$ to replace in the above inequality. By the definition of unnormalized empirical loss on $\mathcal{D}^\prime$:

    \begin{align*}
        &\hat{\mathcal{R}} (\theta_u) - \hat{\mathcal{R}} (\theta^\prime) \\
        &= 
        \sum_{i=1}^{n-1} \ell(f_{\theta_u}(x_i), y_i) + \ell(f_{\theta_u}(x), y)  +  \ell(f_{\theta_u}(x^\prime), y^\prime)
        - \sum_{i=1}^{n-1} \ell(f_{\theta^\prime}(x_i), y_i) - \ell(f_{\theta^\prime}(x), y)  -  \ell(f_{\theta^\prime}(x^\prime), y^\prime) \\
        &=
        \ell(f_{\theta_u}(x), y)  + \ell(f_{\theta_u}(x^\prime), y^\prime)
        - \ell(f_{\theta^\prime}(x), y)  -  \ell(f_{\theta^\prime}(x^\prime), y^\prime),
    \end{align*}

    \noindent where the last equality was derived by the assumption that models are trained until they achieve near-$0$ loss on their corresponding dataset. Therefore, $\sum_{i=1}^{n-1} \ell(f_{\theta_u}(x_i), y_i) = \sum_{i=1}^{n-1} \ell(f_{\theta^\prime}(x_i), y_i) = 0$ since the retrained model has been trained on the remaining samples and the unlearned model has been derived by a single step of gradient descent on the original model, that had been trained on $\mathcal{D}$.

    To further simplify the derived terms above and reaching at our desired inequality, we focus on the term $ - \ell(f_{\theta^\prime}(x), y)$. By adding and decreasing the term $\ell(f_{\theta_o}(x^\prime), y)$ we get:

    \begin{align*}
        - \ell(f_{\theta^\prime}(x), y) &= - \ell(f_{\theta_o}(x^\prime), y) + \ell(f_{\theta_o}(x^\prime), y) - \ell(f_{\theta^\prime}(x), y)  \\
        &\leq 
        - \ell(f_{\theta_o}(x^\prime), y) + \ell(f_{\theta_o}(x^\prime), y)
        - \ell(f_{\theta_o}(x), y) - \langle \nabla \ell(f_{\theta_o}(x), y), \theta^\prime - \theta_o \rangle \\
        &= - \ell(f_{\theta_o}(x^\prime), y) + \ell(f_{\theta_o}(x^\prime), y)
        - \ell(f_{\theta_o}(x), y) \\
        &\leq
        - \ell(f_{\theta_o}(x^\prime), y) + L \delta,
    \end{align*}

    \noindent where the first inequality uses the convexity of the the loss function with respect to the parameters and the third derivations is due to the assumption that the original model achieves a zero loss on its training samples, including $(x,y)$ (hence, $\nabla \ell(f_{\theta_o}(x), y) =0$). The final inequality is due to the Lipschitzness assumption of model $f$ with respect to the inputs.
    
\end{proof}

\section{Related Works (cont.)}
\label{apx:related}

To the best of our knowledge \amun{} is the first work that considers fine-tuning of a model on the adversarial examples with their wrong labels as a method for unlearning a subset of the samples. However, upon reviewing the prior works in unlearning literature, there are several works that their titles might suggest otherwise. Therefore, here we mention a few of these methods and how they differ from our work.

To improve upon fine-tuning on samples in $\Df$ with randomly chosen wrong labels,~\citet{chen2023boundary} use the labels derived from one step of the FGSM attack to choose the new labels for the samples in $\Df$. This method which was presented as \texttt{BS} in our experiments (\S~\ref{sec:results}), does not use the adversarial examples and only uses their labels as the new labels for samples of $\Df$. This corresponds to the dataset $\D_{AdvL}$ in \S~\ref{sec:ablation-finetune}. As our results in Figures~\ref{fig:fine_tune_10} and~\ref{fig:fine_tune_50} show, fine-tuning the trained model on this dataset leads to catastrophic forgetting even when $\Dr$ is available. This is simply due to the fact that the samples in $\D_{AdvL}$ contradict the distribution that the trained models have already learned.

The work by~\citet{setlur2022adversarial} is not an unlearning method, despite what the name suggest. They propose a regularization method that tries to maximize the loss on the adversarial examples of the training samples that are relatively at a higher distance to lower the confidence of the model on those examples. The work by~\citet{zhang2024defensive} proposed a defense method similar to adversarial training for making to unlearned LLMs more robust to jailbreak attacks on the topics that they have unlearned. \citet{lucki2024adversarial} also study the careful application of jailbreak attacks against unlearned models. The work by~\citet{jung2024attack} investigate computing adversarial noise to mask the model parameters. Many of the works with similar titles, use ``adversarial" to refer to minimax optimization~\citep{zeng2021adversarial} or considering a Stackelberg game setting between the source model and the adversary that is trying to extract information~\citep{di2024adversarial}.

\section{Implementation Details}
\label{apx:impl_details}

For all the experiments we train three models on $\D$. For each size of $\D$ ($10\%$ or $50\%$), we use three random subsets and for each subset, we try three different runs of each of the unlearning methods. This leads to a total of $27$ runs of each unlearning method using different initial models and subsets of $\D$ to unlearn. Hyper-parameter tuning of each of the methods is done on a separate random subset of the same size from $\D$, and then the average performance is computed for the other random subsets used as $\Df$. 
For tuning the hyper-parameters of the models, we followed the same range suggested by their authors and what has been used in the prior works for comparisons. Similar to prior works~\cite{liu2024model,fan2023salun}, we performed 10 epochs for each of the unlearning methods, and searched for best learning rate and number of steps for a learning rate scheduler. More specifically, for each unlearning method, we performed a grid search on learning rates within the range of $[10^{-6}, 10^{-1}]$ with an optional scheduler that scales the learning rate by $0.1$ for every $1$ or $5$ steps. For \texttt{SalUn}, whether it is used on its own or in combination with \amun{}, we searched for the masking ratios in the range $[0.1,0.9]$.

The original models are ResNet-18 models trained for 200 epochs with a learning rate initialized at 0.1 and using a scheduler that scales the learning rate by 0.1 every 40 epochs. The retrained models are trained using the same hyper-parameters as the original models. 
For evaluation using RMIA, we trained $128$ separate models such that each sample is included in half of these models. As suggested by the authors, we used Soft-Margin Taylor expansion of Softmax (SM-Taylor-Softmax) with a temperature of $2$ for deriving the confidence values in attacks of RMIA. We used the suggested threshold of $2$ for comparing the ratios in computing the final scores ($\gamma$ value).
For controlling the Lipschitz constant of the ResNet-18 models in~\S~\ref{sec:ablation-robust}, we used the default setting provided by the authors for clipping the spectral norm of all the convolutional and fully-connected layers of the model to $1$. For RMIA evaluations, we trained 128 of these models separately such that each sample appears in exactly half of these models.

\section{\amun{} + SalUn}
\label{apx:amun_salun}

The main idea behind SalUn is to limit the fine-tuning of the model, during unlearning, to only a subset of the parameters of the model, while keeping the rest of them fixed. \citet{fan2023salun} show that this technique helps to preserve the accuracy of the model when fine-tuning the model on $\Df$ with randomly-chosen wrong labels. More specifically, they compute a mask using the following equation:

 \vspace*{-5mm}
 {\begin{equation*}
      \mathbf m_{\mathrm{S}} =  {\bf 1} \left ( \left |  \nabla_{\thetafull} \ell (\thetafull; \Df) \left . \right | \right  | \geq  \gamma \right ),
     \label{eq: sal_map_hard}
 \end{equation*}%
 which, basically, computes the gradient of the loss function for the current parameters with respect to $\Df$, and uses threshold $\gamma$ to filter the ones that matter more to the samples in $\Df$. Note that,  $ {\bf 1}$ is an element-wise indicator function. Then, during fine-tuning of the model on $\Df$ with random labels they use $\mathbf m_{\mathrm{S}}$ to detect the parameters of $\thetafull$ that get updated.

 In our experiments, we try combining this idea with \amun{} for updating a subset of the parameters that might be more relevant to the samples in $\Df$. We refer to this combination as \amun{}$_{SalUn}$ in Tables~\ref{tab:rmia} and~\ref{tab:rmia_forgetonly} in \S~\ref{sec:results} and Tables~\ref{tab:mia} and~\ref{tab:mia_forgetonly} in Appendix~\ref{apx:svc_mia}. As the results show, \amun{}$_{SalUn}$ constantly outperforms \texttt{SalUn} and for the cases that $\Dr$ is not available it also outperforms \amun{}. In the setting where $\Dr$ is accessible, it performs comparable to \amun{}. This is probably due to the fact that when $\Dr$ is not available and \amun{} has access to only the samples $\Df \cup \Da$, SalUn acts as a regularization for not allowing all the parameters of the model that might not be relevant to $\Df$ be updated. In the setting where $\Dr$ is available, involving it in fine-tuning will be a sufficient regularization that preserves models' utility while unlearning $\Df$.

\begin{figure}[t!]
\centering
\begin{subfigure}
    \centering
    \includegraphics[width=.32\linewidth]{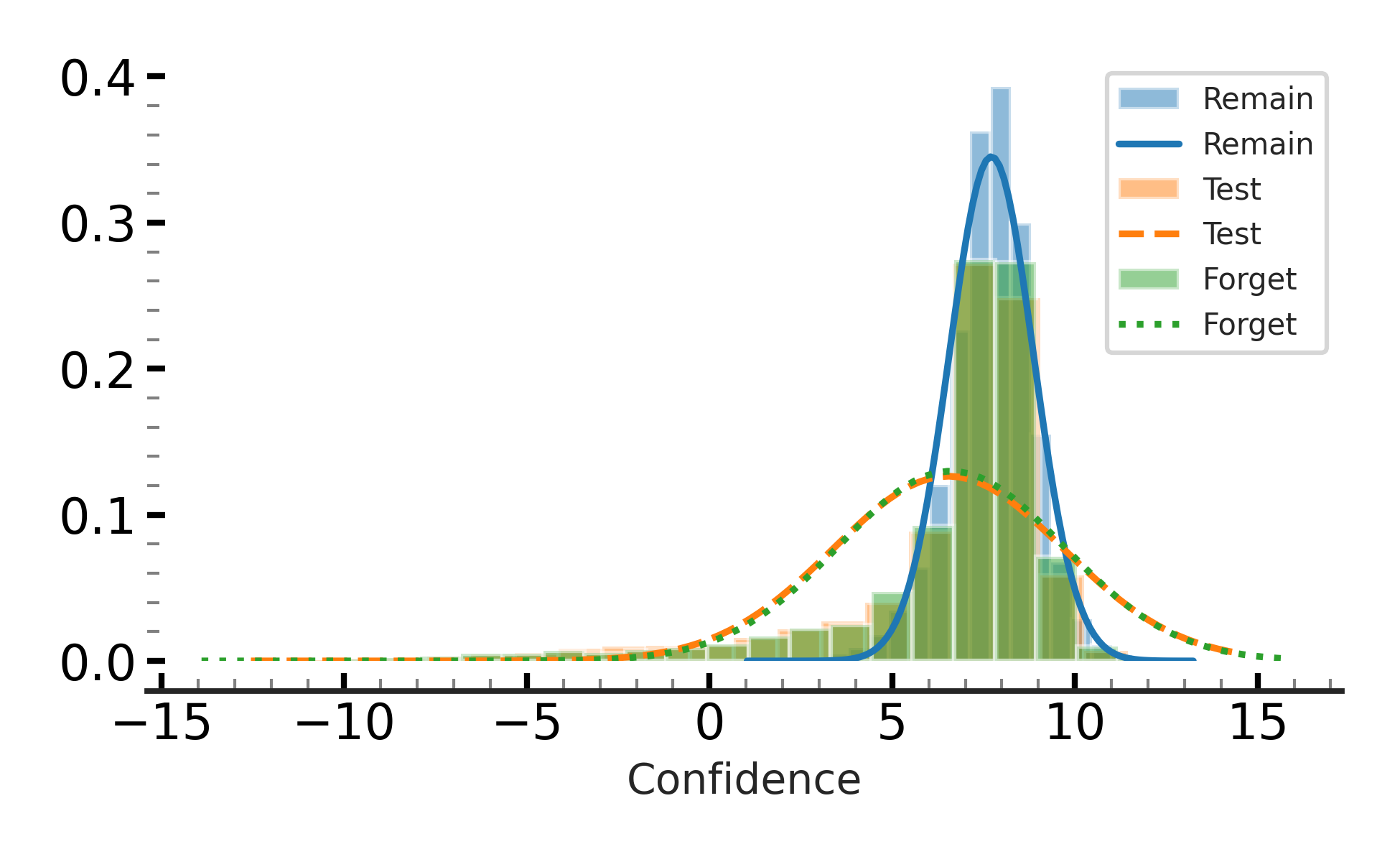}
\end{subfigure}
\begin{subfigure}
    \centering
    \includegraphics[width=.32\linewidth]{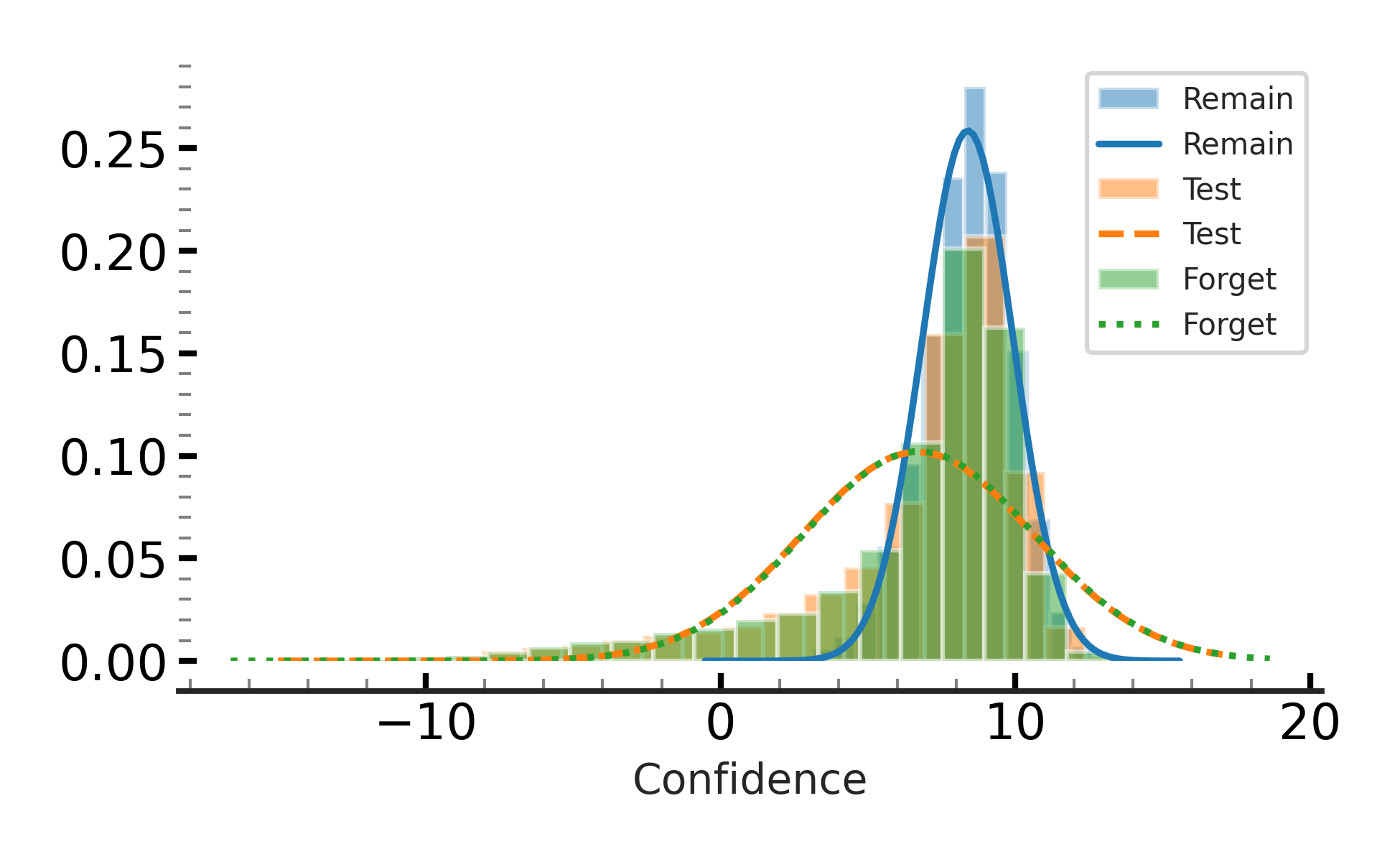}  
\end{subfigure}
\begin{subfigure}
    \centering
    \includegraphics[width=.32\linewidth]{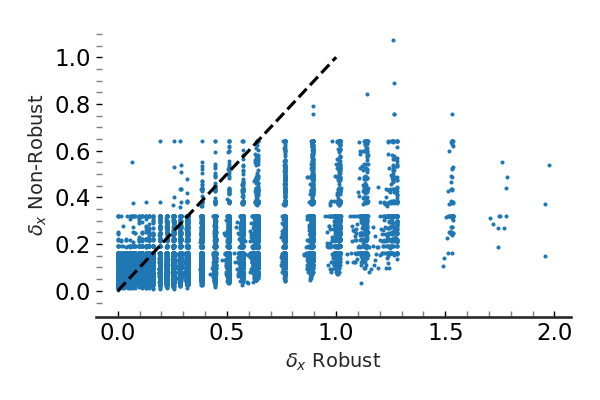}  
\end{subfigure}

\caption{
(left) These two plots show the histogram of confidence values of the retrained model on its predictions for the remaining set (Remain), test set (Test), and forget set (Forget) during the training, when the size of the forget set is $\%10$ (1st plot) and $\%50$ (2nd plot) of the training set. It also shows the Gaussian distributions fitted to each histogram. As the plots show the models perform similarly on the forget set and test set because to the retrained model they are unseen data from the same distribution. (right) This plot compares the $\delta_x$ value in definition~\ref{def:attack} for adversarial examples generated on the original ResNet-18 models (x-axis) and clipped ResNet-18 models (y-axis). The dashed line shows $x=y$ line and more than $97\%$ of the values fall bellow this line. } 
\vspace{-5mm}
\label{fig:retrain_conf}
\end{figure}

\textbf{}
\section{Ablation Study (cont.)}
\label{apx:ablation_all}

In this section, we further discuss the ablation studies that were mentioned in \S~\ref{sec:ablation}. We also present other ablation studies on using transferred adversarial examples (Appendix~\ref{apx:transfer_attack}) and weaker adversarial attacks (Appendix~\ref{apx:weak_attack}) in \amun{}.

\subsection{Empirical Behavior of Retrained Models}
\label{apx:confidence}

As discussed in \S~\ref{sec:motivation}, assuming the $\Dt$ and $\D$ come from the same distributions, we expect the prediction confidences of the models retrained on $\Dr$ to be similar on $\Df$ and $\Dt$, because both of these sets are considered unseen samples that belong the the same data distribution.  Figure~\ref{fig:retrain_conf} (left) shows the confidence scores for a ResNet-18 model that has been retrained on $\D - \Df$, where $\D$ is the training set of CIFAR-10 and the size of $\Df$ (randomly chosen from $\D$) is $10\%$ and $50\%$ of the size of $\D$ for the left and right sub-figures, respectively. To derive the confidence values, we use the following scaling on the logit values:

\begin{align*}
    \phi(f(x)_y) = \mathrm{log} \left( \frac{f(x)_y}{1-f(x)_y} \right),
\end{align*}

\noindent where $f(x)_y$ is the predicted probability for the correct class. This scaling has been used by~\citet{carlini2022membership} to transform the the prediction probabilities such that they can be better approximated with a normal distribution, which are indeed used by some of the SOTA MIA methods for predicting training samples from the test samples~\cite{carlini2022membership}. Figure~\ref{fig:retrain_conf} (left) shows these fitted normal distribution as well, which perfectly match for $\Df$ and $\Dt$.

\subsubsection{Confidence Values in Unlearned Models}
\label{apx:confidence_after}

In this section, we investigate the confidence values of the model, before and after using \amun{} for unlearning a subset of $10\%$ or $25\%$ of the training samples. For the original model (before unlearning), we expect the distribution of confidence values of samples in $\Df$ to be similar to those of the samples in $\Dr$ because they were both used as the training data and the model has used them similarly during training. However, this distribution is different for the test samples ($\Dt$), as the model has not seen them during the training phase. After unlearning, as discussed in section~\ref{sec:motivation}, we expect the distribution of confidence values for $\Df$ and $\Dt$ to become more similar so that MIAs cannot distinguish them from each other. As Figure~\ref{fig:unlearn_conf} shows, for both unlearning $10\%$ (two leftmost subplots) and $50\%$ (two rightmost subplots), we observe the same behavior. Fur the original models (1st and 3rd subplot), the distribution for $\Df$ and $\Dr$ mathces exactly, but after using \amun{} (2nd and 4th subplot) the distribution for $\Df$ shifts toward that of $\Dt$.

\begin{figure}[t!]
\centering
\begin{subfigure}
    \centering
    \includegraphics[width=.24\linewidth]{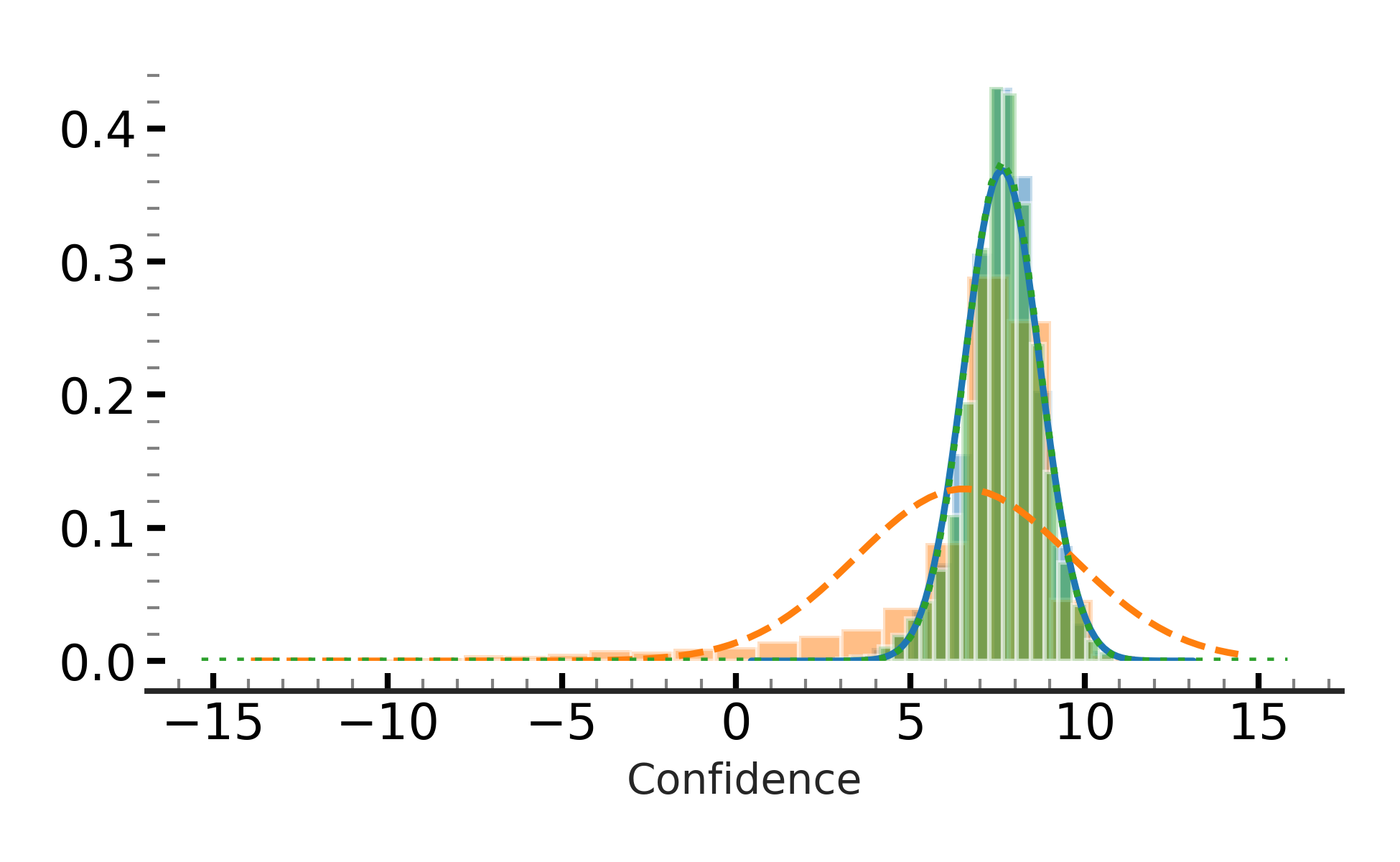}
    \includegraphics[width=.24\linewidth]{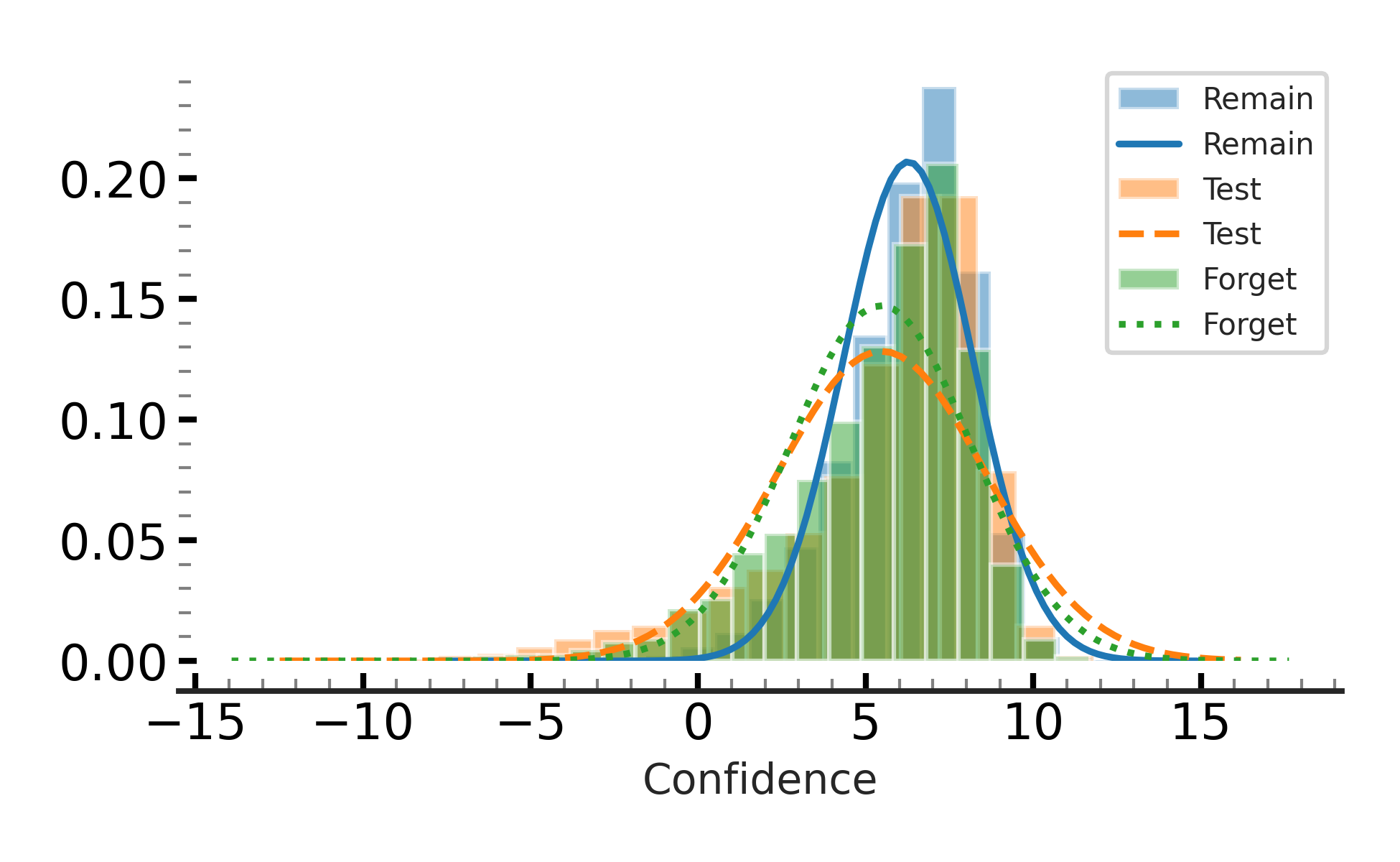}  
\end{subfigure}
~ 
\begin{subfigure}
    \centering
    \includegraphics[width=.24\linewidth]{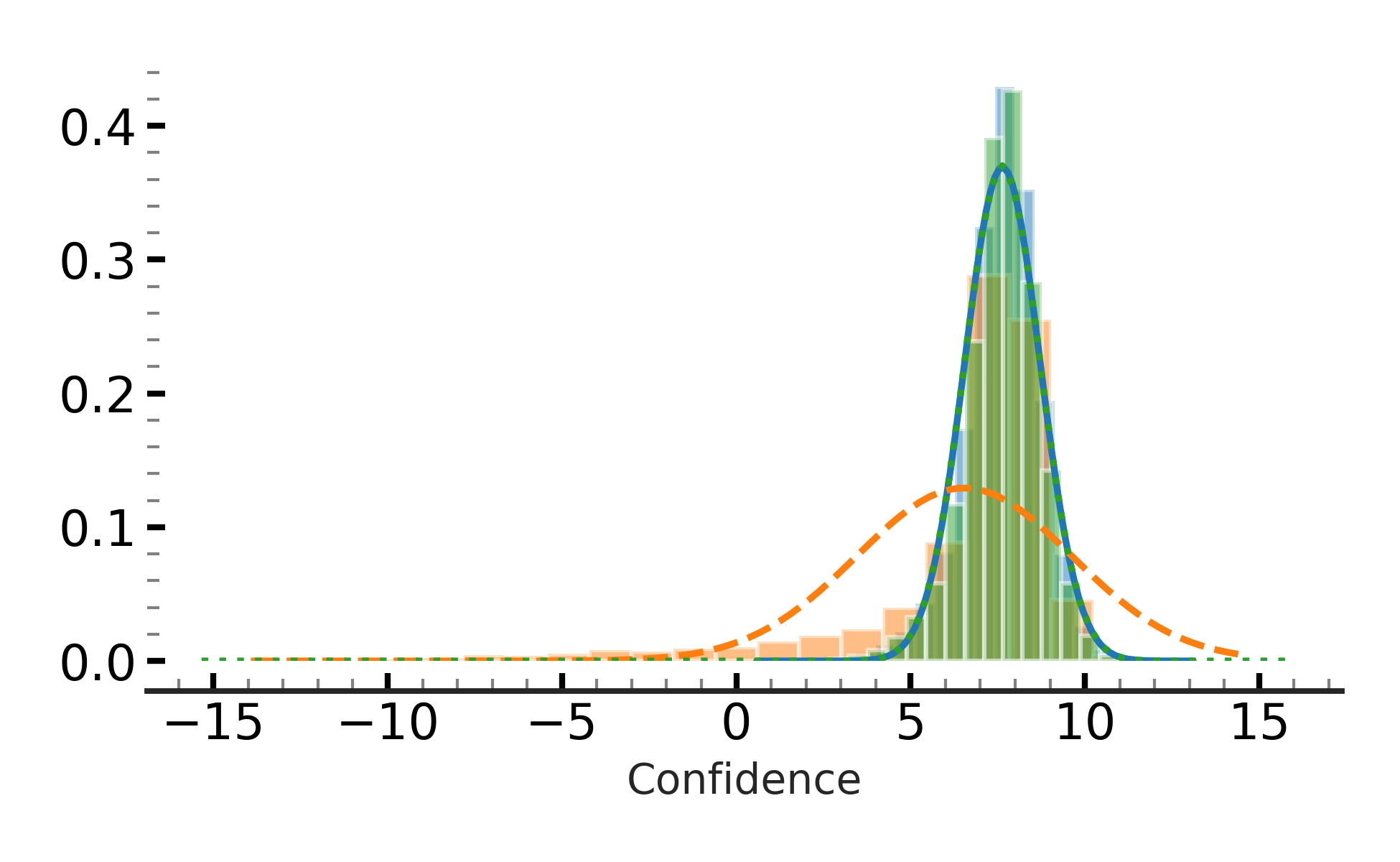}  
    \includegraphics[width=.24\linewidth]{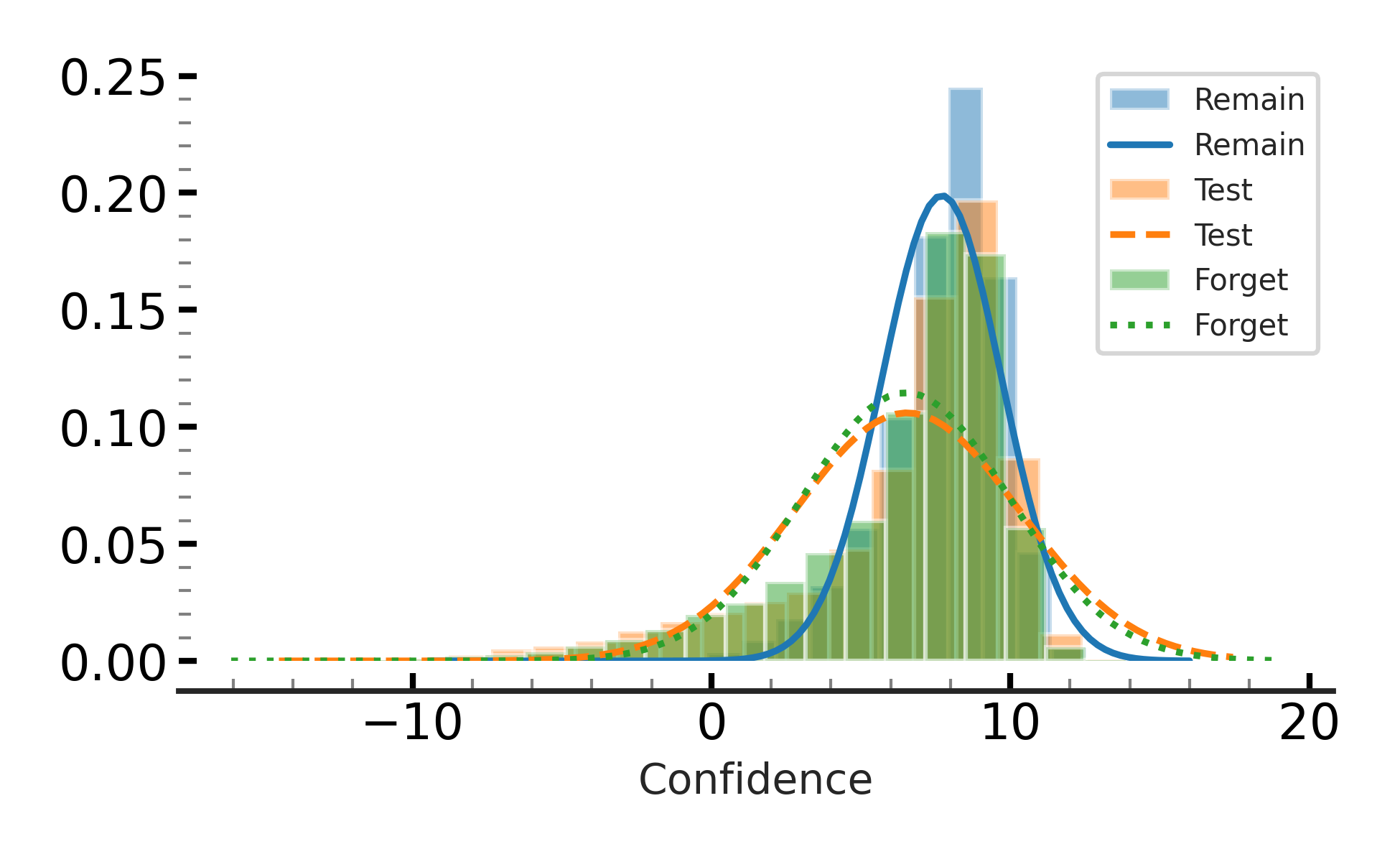}  
\end{subfigure}

\caption{The two left-most subplots show the confidence values before and after unlearning (using \amun{}) of $10\%$ of the training samples. The two right-most subplots show these confidence values for unlearning $50\%$ of the training samples. In both cases, the confidence values of samples in $\Df$ are similar to those of $\Dr$ and their fitted Gaussian distribution matches as expected. After using \amun{} for unlearning the samples in $\Df$, the confidence values on this set gets more similar to the test (unseen) samples.} 
\vspace{-5mm}
\label{fig:unlearn_conf}
\end{figure}

\subsection{Adversarially Robust Models (cont.)}
\label{apx:ablation_robust}

As discussed in \S~\ref{sec:ablation-robust}, we also evaluatee the effectiveness of \amun{} when the trained model is adversarially robust. For this experiment, we used the ResNet-18 models with $1$-Lipschitz convolutional and fully-connected layers, which are shown to be significantly more robust than the original ResNet-18 models. In Table~\ref{tab:rmia-clipped-remFalse}, we showed the results for unlearning $10\%$ and $50\%$ of the samples from the robust ResNet-18 models trained on CIFAR-10, in the case where $\Dr$ is not accessible. In Table~\ref{tab:rmia-clipped-remTrue}, we showed the corresponding results when the unlearning methods have access to $\Dr$. As the results show, similar to the results discussed in \S~\ref{sec:ablation-robust}, \amun{} effectively unlearns $\Df$ for either of the sizes of the this set.

\begin{table}[th!]
\begin{center}
\begin{small}
\begin{sc}
\resizebox{0.66\columnwidth}{!}{
\begin{tabular}{@{} l  c c c | c c c @{}}
 \toprule


& \multicolumn{3}{@{}c}{\textbf{Random Forget ($10\%$)}} & \multicolumn{3}{@{}c}{\textbf{Random Forget ($50\%$)}} \\\addlinespace[0.3em]

 \multicolumn{1}{c}{\scriptsize \textbf{}} & 
 \multicolumn{1}{c}{\scriptsize FT AUC} & 
 \multicolumn{1}{c}{\scriptsize FR AUC} & 
 \multicolumn{1}{c}{\scriptsize Test Acc} & 
 \multicolumn{1}{c}{\scriptsize FT AUC} &
 \multicolumn{1}{c}{\scriptsize FR AUC} & 
 \multicolumn{1}{c}{\scriptsize Test Acc} 
 \\\addlinespace[0.3em]

 \cmidrule(r){2-4}
 \cmidrule(r){5-7}

 Retrain & $49.95$ {\tiny $\pm 0.24$} & $54.08 $ {\tiny $\pm 0.16$} & $89.01$ {\tiny $\pm 0.21$} & 
 
 $50.19$ {\tiny $\pm 0.15$}  & $55.61$ {\tiny $\pm 0.05$} & $85.76$ {\tiny $\pm 0.41$}
 
 \\\addlinespace[0.3em]
  \cmidrule(r){1-7}

 \textbf{Amun} & $49.12$ {\tiny $\pm 0.19$} & $53.60$ {\tiny $\pm 0.31$} & $86.94$ {\tiny $\pm 0.56$} & 
 
 $49.41$ {\tiny $\pm 0.25$} & $54.22$ {\tiny $\pm 0.16$} & $87.38$ {\tiny $\pm 0.39$}
 \\\addlinespace[0.3em]

\bottomrule
\end{tabular}
}
\end{sc}
\end{small}
\end{center}
\caption{\footnotesize {\bf Unlearning on adversarially robust models.} Evaluating the effectiveness of \amun{} in unlearning $10\%$ and $50\%$ of the training samples when the models are adversarially robust and we have access to $\Dr$. For this experiment we use models with controlled Lipschitz constant which makes them provably and empirically more robust to adversarial examples. \vspace{-3mm}}
\label{tab:rmia-clipped-remTrue}

\end{table}

We also evaluated \amun{} for unlearning in models that are adversarially trained. we performed our analysis on ResNet-18 models trained using TRADES loss~\cite{zhang2019theoretically} on CIFAR-10. We performed the experiments for unlearning 10\% of the dataset in both cases where $\Dr$ is accessible and not. As the results in Table~\ref{} show, in both settings \amun{} is effective in unlearning the forget samples and achieving a low gap with the retrained models. This gap is obviously smaller when there is access to $\Dr$.

\begin{table*}[th!]
\begin{center}
\begin{small}
\begin{sc}
\begin{tabular}{@{} l  c c  c  c c @{}}
 \toprule



 \multicolumn{1}{c}{\scriptsize \textbf{}} & 
 \multicolumn{1}{c}{\scriptsize Unlearn Acc} & 
 \multicolumn{1}{c}{\scriptsize Retain Acc} & 
 \multicolumn{1}{c}{\scriptsize Test Acc} & 
 \multicolumn{1}{c}{\scriptsize FT AUC} & 
 \multicolumn{1}{c}{\scriptsize Avg. Gap} 
 \\\addlinespace[0.3em]

 \cmidrule(r){2-6}



 
 

 Retrain & $82.33$ {\tiny $\pm 0.39$} & $94.22 $ {\tiny $\pm 0.21$} & $81.72$ {\tiny $\pm 0.36$} & $50.04$ {\tiny $\pm 0.34$} & $0.00$ 
 \\\addlinespace[0.3em]

 \textbf{Amun}$_{\, \mathrm{With} \, \Dr}$  & $82.65$  { \tiny $\pm 0.62$ }  & $94.33$  { \tiny $\pm 0.84$ }  & $84.99$  { \tiny $\pm 0.91$ }  & $47.18$  { \tiny $\pm 0.50$ }  & $1.02$  { \tiny $\pm 0.18$ } 
 \\\addlinespace[0.3em]

 \textbf{Amun}$_{\, \mathrm{No} \, \Dr}$  & $81.38$  { \tiny $\pm 0.10$ }  & $87.45$  { \tiny $\pm 0.54$ }  & $79.74$  { \tiny $\pm 0.31$ }  & $54.61$  { \tiny $\pm 0.23$ }  & $3.57$  { \tiny $\pm 0.24$ } 
 \\\addlinespace[0.3em]

 \bottomrule
\end{tabular}
\end{sc}
\end{small}
\end{center}
\caption{{\bf Unlearning with access to $\Dr$.} Evaluating \amun{} when applied to ResNet-18 models trained using adversarial training. TRADES loss is used to train the models, and the unlearning is done on $10\%$ of CIFAR-10 Dataset ($\D$). Avg. Gap is used for evaluation (lower is better). The result has been reported in two cases: with and without access to $\Dr$. As the results show, \amun{} is effective in both cases, with slight degradation in the more difficult setting of no access to $\Dr$.}
\label{tab:trades_results}

\end{table*}

\subsection{Fine-tuning on Adversarial Examples (cont.)}
\label{apx:ablation-finetune}

As explained in \S~\ref{sec:ablation-finetune}, we evaluate the effect of fine-tuning on test accuracy of a ResNet-18 model that is trained on CIFAR-10, when $\Da$ is substituted with other datasets that vary in the choice of samples or their labels (see \S~\ref{sec:ablation-finetune} for details). In Figure~\ref{fig:fine_tune_10} we presented the results when $\Df$ contains $10\%$ of the samples in $\D$. We also present the results for the case where  $\Df$ contains $50\%$ of the samples in $\D$ in Figure~\ref{fig:fine_tune_50}. As the figure shows, even for the case where we fine-tune the trained models on only $\Da$ which contains the adversarial examples corresponding to $50\%$ of the samples in $\D$ (right-most sub-figure), there is no significant loss in models' accuracy. This is due to the fact that the samples in $\Da$, in contrast to the other constructed datasets, belong to the natural distribution learned by the trained model. To generate the results in both Figures~\ref{fig:fine_tune_10} and~\ref{fig:fine_tune_50}, we fine-tuned the trained ResNet-18 models on all the datasets (see \S~\ref{sec:ablation-finetune} for details) for 20 epochs. We used a learning rate of $0.01$ with a scheduler that scales the learning rate by $0.1$ every 5 epochs.

\begin{figure*}[t!]
\centering
\includegraphics[width=.98\linewidth]{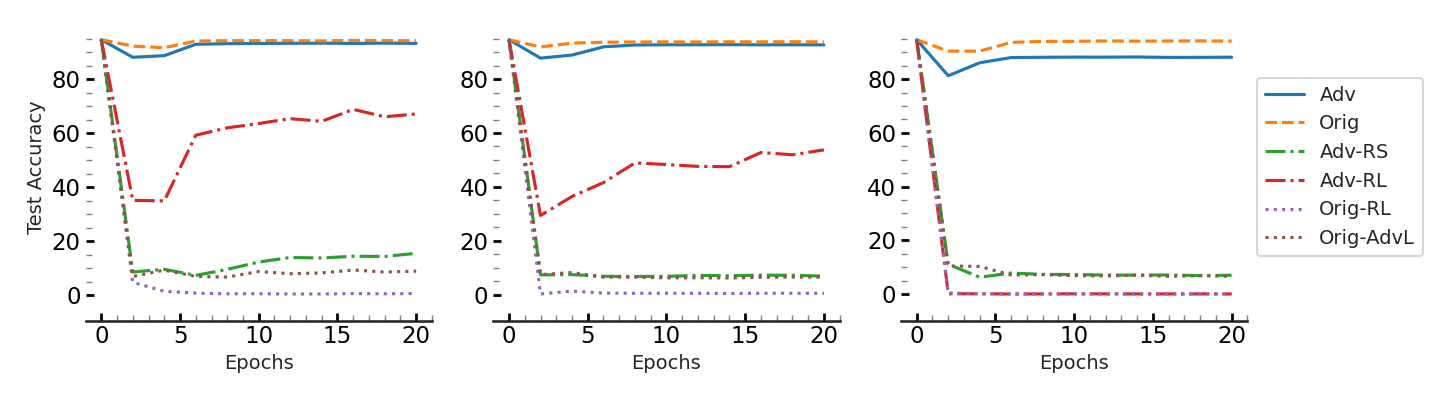}
\caption{This figure shows the effect of fine-tuning on test accuracy of a ResNet-18 model that is trained on CIFAR-10, when the dataset for fine-tuning changes (see \S~\ref{sec:ablation} for details). Let $\Df$ contain $50\%$ of the samples in $\D$ and $\Da$ be the set of adversarial examples constructed using Algorithm~\ref{alg:advset}. \texttt{Adv}, from the left sub-figure to right one, shows the results when $\D \cup \Da$, $\Df \cup \Da$, and $\Da$ is used for fine-tuning the model, respectively. \texttt{Orig}, \texttt{Adv-RS}, \texttt{Adv-RL}, \texttt{Orig-RL}, and \texttt{Orig-AdvL} shows the results when $\Da$ for each of these sub-figures is replace by $\Df$, $\Da_{RS}$, $\Da_{RL}$, $\D_{RL}$, and $\D_{AdvL}$, accordingly. As the figure shows, the specific use of adversarial examples with the mis-predicted labels matters in keeping the model's test accuracy because $\Da$, in contrast to the other constructed datasets belong to the natural distribution learned by the trained model.} 
\vspace{-5mm}
\label{fig:fine_tune_50}
\end{figure*}

\subsection{Transferred Adversarial Examples}
\label{apx:transfer_attack}

One of the intriguing properties of adversarial attacks is their transferability to other models~\cite{papernot2016transferability,liu2016delving}; Adversarial examples generated on a trained model (source model) mostly transfer successfully to other models (target models). This success rate of the transferred adversarial examples increases if the source model and target model have the same architecture~\cite{papernot2016transferability}. There are other studies that can be used to increase the success rate of this type of attack~\cite{zhao2021success,zhang2022improving,chen2023rethinking,ebrahimpour2024lotos}. In this section, we are interested to see if using the the adversarial examples generated using Algorithm~\ref{alg:advset} for a given model trained on some dataset $\D$ can be used as the $\Da$ dataset for unlearning a portion of $\D$ from a separately trained model. The advantage of using adversarial examples generated for another model is saving the computation cost for other trained models. For this purpose, we train three ResNet-18 models separately on CIFAR-10, we generate the adversarial examples for each of these models using Algorithm~\ref{alg:advset}. We use \amun{} for unlearning $10\%$ and $50\%$ of CIFAR-10 from either of these models, but instead of their adversarial samples, we use the ones derived from the other models. The results in Table~\ref{tab:transfer} shows that using transferred adversarial examples leads to lower performance, specially for the case where there is no access to $\Dr$. All the values for test accuracy are also lower compared to using adversarial examples from the model itself because these adversarial examples from the other models do not all belong to the natural distribution of the model and they do not even always transfer to the other models. Still the results are comparable to the prior SOTA methods in unlearning, and even in the case of no access to $\Dr$ outperforms all prior methods.

\begin{table*}[th!]
\begin{center}
\begin{small}
\begin{sc}
\resizebox{0.98\textwidth}{!}{
\begin{tabular}{@{} l  c c c c c c | c c c c c c @{}}
 \toprule


& \multicolumn{6}{@{}c}{\textbf{With access to $\Dr$}} & \multicolumn{6}{@{}c}{\textbf{No access to $\Dr$}} \\\addlinespace[0.3em]

& \multicolumn{3}{@{}c}{\textbf{Random Forget ($10\%$)}} & \multicolumn{3}{@{}c}{\textbf{Random Forget ($50\%$)}} 
& \multicolumn{3}{@{}c}{\textbf{Random Forget ($10\%$)}} & \multicolumn{3}{@{}c}{\textbf{Random Forget ($50\%$)}} \\\addlinespace[0.3em]

 \multicolumn{1}{c}{\scriptsize \textbf{}} & 
 \multicolumn{1}{c}{\scriptsize Test Acc} & 
 \multicolumn{1}{c}{\scriptsize FT AUC} & 
 \multicolumn{1}{c}{\scriptsize Avg. Gap} & 
 \multicolumn{1}{c}{\scriptsize Test AUC} &
 \multicolumn{1}{c}{\scriptsize FT AUC} & 
 \multicolumn{1}{c}{\scriptsize Avg. Gap} &

  \multicolumn{1}{c}{\scriptsize Test Acc} & 
 \multicolumn{1}{c}{\scriptsize FT AUC} & 
 \multicolumn{1}{c}{\scriptsize Avg. Gap} & 
 \multicolumn{1}{c}{\scriptsize Test AUC} &
 \multicolumn{1}{c}{\scriptsize FT AUC} & 
 \multicolumn{1}{c}{\scriptsize Avg. Gap} 
 \\\addlinespace[0.3em]

 \cmidrule(r){2-4}
 \cmidrule(r){5-7}
 \cmidrule(r){8-10}
 \cmidrule(r){11-13}



 \textbf{Self} & $93.45$ {\tiny $\pm 0.22$} & $50.18 $ {\tiny $\pm 0.36$} & $0.62$ { \tiny $\pm 0.05$ } & 
 
 $92.39$ {\tiny $\pm 0.04$}  & $49.99$ {\tiny $\pm 0.18$} & $0.33$ { \tiny $\pm 0.03$ } &

 $91.67$ {\tiny $\pm 0.04$}  & $52.24$ {\tiny $\pm 0.23$} & $1.94$ { \tiny $\pm 0.13$ } &

 $89.43$ {\tiny $\pm 0.19$}  & $52.60$ {\tiny $\pm 0.22$} & $2.51$ { \tiny $\pm 0.09$ } 
 
 \\\addlinespace[0.3em]

 \textbf{Others} & $92.64$  { \tiny $\pm 0.09$ }  & $48.70$  { \tiny $\pm 0.59$ }  & $1.57$  { \tiny $\pm 0.12$ }  & 
 
$91.49$  { \tiny $\pm 0.03$ }  & $47.36$  { \tiny $\pm 0.63$ }  & $1.15$  { \tiny $\pm 0.23$ }  &

 $90.56$  { \tiny $\pm 0.28$ }  & $48.29$  { \tiny $\pm 0.22$ }  & $3.07$  { \tiny $\pm 0.15$ }  &
 
$83.61$  { \tiny $\pm 0.45$ }  & $51.11$  { \tiny $\pm 0.04$ }  & $6.70$  { \tiny $\pm 0.33$ }

 \\\addlinespace[0.3em]

 

 
 
 \bottomrule
\end{tabular}
}
\end{sc}
\end{small}
\end{center}
\caption{\footnotesize {\bf Transferred adversarial examples.} Comparing the effectiveness of unlearning when instead of using adversarial examples of the model, we use adversarial examples generated using Algorithm~\ref{alg:advset} on separately trained models with the same architecture. As the results show, relying on transferred adversarial examples in \amun{} leads to worse results, specially for test accuracy because the adversarial examples do not necessary belong to the natural distribution learned by the model. However, even by using these transferred adversarial examples \amun{} outperforms prior SOTA unlearning methods, specially when there is no access to $\Dr$.\vspace{-3mm}}
\label{tab:transfer}

\end{table*}

\begin{figure*}[t!]
\centering
\includegraphics[width=.98\linewidth]{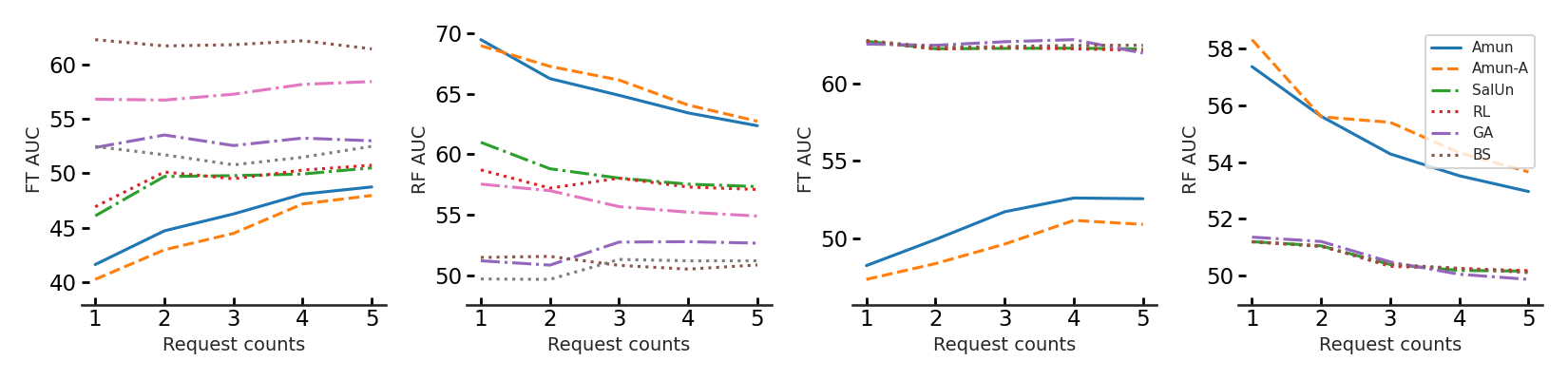}
\caption{This figure shows both \texttt{FT AUC} and \texttt{RF AUC} components of the plots presented in Figure~\ref{fig:adaptive}. The two left-most sub-figures show these values along the number of unlearning requests for the case where there is access to $\Dr$ and the two right-most ones show these values when there is no access to $\Dr$. } 
\vspace{-5mm}
\label{fig:adaptive_FT_RF}
\end{figure*}

\subsection{Weak Attacks}
\label{apx:weak_attack}

In this section we evaluate the effectiveness of using weaker attacks in Algorithm~\ref{alg:advset}. For this purpose, we perform the unlearning on a ResNet-18 model trained on CIFAR-10 in all unlearning settings mentioned in \S~\ref{sec:evaluation}, and compare the results with the default choice of PGD-50 in \amun{}. The weaker attack that we use is a variation of FFGSM~\cite{wong2020fast}, which itself is a variant of FGSM~\cite{goodfellow2014explaining}. FGSM takes steps toward the gradient sign at a given sample to find adversarial samples. FFGSM takes a small step toward a random direction first, and then proceeds with FGSM. To adapt these method to the format of Algorithm~\ref{alg:advset} we start with FGSM attack; we find the gradient sign and start to move toward that direction in steps of size $\epsilon$ until we find an adversarial example. If the adversarial example is not found after a few iteration of the \texttt{While} loop, we restart the value of $\epsilon$ and add a small random perturbation before the next round of FGSM attack and the \texttt{While} loop. We continue this procedure to find an adversarial sample. After deriving a new set of adversarial examples using this methods, we performed a separate round hyper-parameter tuning for unlearning with the new attack to have a fair comparison. It is notable to mention that this leads to a much faster attack because we only compute the gradient once for each round round of FGSM (at the beginning or after each addition of random perturbation and restarting FGSM). Table~\ref{tab:fgsm} shows the comparison of the results with the original version of \amun{} that uses PGD-50. As the results show, using this weaker attack leads to worse results; however, they still outperform prior SOTA methods in unlearning, specially in the setting where there is no access to $\Dr$ and the size of $\Df$ is $50\%$ of $\D$.

For each image in CIFAR-10, Figure~\ref{fig:fgsm_ratio} shows $\delta_x$ (see Definition~\ref{def:attack}) for the adversarial examples that Algorithm~\ref{alg:advset} finds using PGD-50 ($x$-axis) and FFGSM ($y$-axis). The dashed line shows the $x=y$ line for the reference. As the figure shows $\delta_x$ is much smaller for PGD-50. This value is smaller for FFGSM for less than $4\%$ of the images, but still even for those images, the value of $\delta_x$ for PGD-50 is very small, compared to the range of values that are required for FFGSM in many cases. This, we believe, is the main reason behind worse performance when using FFGSM. However, still note that the adversarial examples that are found using FFGSM belong to the natural distribution of the trained model and therefore fine-tuning the model on these samples does not lead to noticable deterioration of the test accuracy, while achieving reasonable \texttt{FT AUC} score. Indeed this larger distance of the adversarial examples with the original samples in $\Df$, leads to better performance of \amun{} when it does not include $\Df$ when fine-tuning the model, because the difference in the predicted logits compared to the $\delta_x$ leads to under-estimation of the local Lipschitz constant and therefore, the model is able to fit perfectly to both the original samples and its corresponding adversarial sample without changing much. This consequently leads to a larger value of \texttt{FT AUC} score.

\begin{table*}[th!]
\begin{center}
\begin{small}
\begin{sc}
\resizebox{0.98\textwidth}{!}{
\begin{tabular}{@{} l  c c c c c c | c c c c c c @{}}
 \toprule


& \multicolumn{6}{@{}c}{\textbf{With access to $\Dr$}} & \multicolumn{6}{@{}c}{\textbf{No access to $\Dr$}} \\\addlinespace[0.3em]

& \multicolumn{3}{@{}c}{\textbf{Random Forget ($10\%$)}} & \multicolumn{3}{@{}c}{\textbf{Random Forget ($50\%$)}} 
& \multicolumn{3}{@{}c}{\textbf{Random Forget ($10\%$)}} & \multicolumn{3}{@{}c}{\textbf{Random Forget ($50\%$)}} \\\addlinespace[0.3em]

 \multicolumn{1}{c}{\scriptsize \textbf{}} & 
 \multicolumn{1}{c}{\scriptsize Test Acc} & 
 \multicolumn{1}{c}{\scriptsize FT AUC} & 
 \multicolumn{1}{c}{\scriptsize Avg. Gap} & 
 \multicolumn{1}{c}{\scriptsize Test AUC} &
 \multicolumn{1}{c}{\scriptsize FT AUC} & 
 \multicolumn{1}{c}{\scriptsize Avg. Gap} &

  \multicolumn{1}{c}{\scriptsize Test Acc} & 
 \multicolumn{1}{c}{\scriptsize FT AUC} & 
 \multicolumn{1}{c}{\scriptsize Avg. Gap} & 
 \multicolumn{1}{c}{\scriptsize Test AUC} &
 \multicolumn{1}{c}{\scriptsize FT AUC} & 
 \multicolumn{1}{c}{\scriptsize Avg. Gap} 
 \\\addlinespace[0.3em]

 \cmidrule(r){2-4}
 \cmidrule(r){5-7}
 \cmidrule(r){8-10}
 \cmidrule(r){11-13}



 \textbf{PGD-50} & $93.45$ {\tiny $\pm 0.22$} & $50.18 $ {\tiny $\pm 0.36$} & $0.62$ { \tiny $\pm 0.05$ } & 
 
 $92.39$ {\tiny $\pm 0.04$}  & $49.99$ {\tiny $\pm 0.18$} & $0.33$ { \tiny $\pm 0.03$ } &

 $91.67$ {\tiny $\pm 0.04$}  & $52.24$ {\tiny $\pm 0.23$} & $1.94$ { \tiny $\pm 0.13$ } &

 $89.43$ {\tiny $\pm 0.19$}  & $52.60$ {\tiny $\pm 0.22$} & $2.51$ { \tiny $\pm 0.09$ } 
 
 \\\addlinespace[0.3em]

 \textbf{FGSM} & $93.87$  { \tiny $\pm 0.16$ }  & $50.64$  { \tiny $\pm 0.51$ }  & $0.92$  { \tiny $\pm 0.25$ }  & 
 
 $89.41$  { \tiny $\pm 1.01$ }  & $50.93$  { \tiny $\pm 0.46$ }  & $1.81$  { \tiny $\pm 0.77$ } &

 $92.14$  { \tiny $\pm 0.28$ }  & $56.58$  { \tiny $\pm 1.05$ }  & $3.46$  { \tiny $\pm 0.36$ } &
 
 $90.12$  { \tiny $\pm 0.28$ }  & $54.54$  { \tiny $\pm 0.47$ }  & $3.29$  { \tiny $\pm 0.10$ }

 \\\addlinespace[0.3em]

 

 
 
 \bottomrule
\end{tabular}
}
\end{sc}
\end{small}
\end{center}
\caption{\footnotesize {\bf Using weaker attacks.} Comparing the effectiveness of unlearning when PGD-10 in Algorithm~\ref{alg:advset} is replaced with a variant of FGSM attack, which is considered to be significantly weaker and leads to finding adversarial examples at a much higher distance to the original samples. We evaluate unlearning $10\%$ and $50\%$ of the training samples in CIFAR-10 from a trained ResNet-18 model. As the results show, in both settings of unlearning (with access to $\Dr$ and no access to $\Dr$), using the weaker attack does not perform as well as the original method. However, it still outperforms prior SOTA unlearning methods.\vspace{-3mm}}
\label{tab:fgsm}

\end{table*}

\section{Comparison Using Prior Evaluation Methods}
\label{apx:svc_mia}

In this section we perform similar comparisons to what we presented in section~\ref{sec:results}, but instead of \texttt{FT AUC}, we use the same MIA used by prior SOTA methods in unlearning for evaluations. As mentioned in section~\ref{sec:metrics}, we refer to the score derived by this MIA as \texttt{MIS}.

\begin{table*}[th!]
\begin{center}
\begin{small}
\begin{sc}
\resizebox{\textwidth}{!}{
\begin{tabular}{@{} l  c c  c  c c | c c c c c @{}}
 \toprule


& \multicolumn{5}{@{}c}{\textbf{Random Forget ($10\%$)}} & \multicolumn{5}{@{}c}{\textbf{Random Forget ($50\%$)}} \\\addlinespace[0.3em]

 \multicolumn{1}{c}{\scriptsize \textbf{}} & 
 \multicolumn{1}{c}{\scriptsize Unlearn Acc} & 
 \multicolumn{1}{c}{\scriptsize Retain Acc} & 
 \multicolumn{1}{c}{\scriptsize Test Acc} & 
 \multicolumn{1}{c}{\scriptsize MIS} & 
 \multicolumn{1}{c}{\scriptsize Avg. Gap} & 
 \multicolumn{1}{c}{\scriptsize Unlearn Acc} &
 \multicolumn{1}{c}{\scriptsize Retain Acc} & 
 \multicolumn{1}{c}{\scriptsize Test Acc} & 
 \multicolumn{1}{c}{\scriptsize  MIS} & 
 \multicolumn{1}{c}{\scriptsize Avg. Gap} 
 \\\addlinespace[0.3em]

 \cmidrule(r){2-6}
 \cmidrule(r){7-11}



 Retrain & $94.49$ {\tiny $\pm 0.20$} & $100.0 $ {\tiny $\pm 0.00$} & $94.33$ {\tiny $\pm 0.18$} & $12.53$ {\tiny $\pm 0.32$} & $0.00$  
 & 
 
 $92.09$ {\tiny $\pm 0.37$}  & $100.0$ {\tiny $\pm 0.00$} & $91.85$ {\tiny $\pm 0.33$} & $16.78$ {\tiny $\pm 0.37$} & $0.00$
 \\\addlinespace[0.3em]
  \cmidrule(r){1-11}

 FT  & $95.16$ { \tiny $\pm 0.29$ }  & $96.64$ { \tiny $\pm 0.25$ }  & $92.21$ { \tiny $\pm 0.27$ }  & $11.33$ { \tiny $\pm 0.35$ }  & $1.84$ { \tiny $\pm 0.10$ } 
 & 
 
 $94.24$ { \tiny $\pm 0.30$ }  & $95.22$ { \tiny $\pm 0.31$ }  & $91.21$ { \tiny $\pm 0.33$ }  & $12.10$ { \tiny $\pm 0.72$ }  & $3.06$ { \tiny $\pm 0.24$ } 
 \\\addlinespace[0.3em]
 
 RL & $99.22$ { \tiny $\pm 0.19$ }  & $99.99$ { \tiny $\pm 0.01$ }  & $94.10$ { \tiny $\pm 0.11$ }  & $10.94$ { \tiny $\pm 0.45$ }  & $1.64$ { \tiny $\pm 0.19$ } 
 & 
 
 $92.98$ { \tiny $\pm 1.07$ }  & $94.83$ { \tiny $\pm 1.04$ }  & $89.19$ { \tiny $\pm 0.74$ }  & $12.48$ { \tiny $\pm 0.90$ }  & $3.29$ { \tiny $\pm 0.04$ } 
 \\\addlinespace[0.3em]

  GA & $98.94$ { \tiny $\pm 1.39$ }  & $99.22$ { \tiny $\pm 1.31$ }  & $93.39$ { \tiny $\pm 1.18$ }  & $4.21$ { \tiny $\pm 5.25$ }  & $3.62$ { \tiny $\pm 1.04$ } 
  & 
 
 $99.94$ { \tiny $\pm 0.09$ }  & $99.95$ { \tiny $\pm 0.08$ }  & $94.36$ { \tiny $\pm 0.31$ }  & $0.62$ { \tiny $\pm 0.30$ }  & $6.64$ { \tiny $\pm 0.15$ } 
 \\\addlinespace[0.3em]

 BS & $99.14$ { \tiny $\pm 0.31$ }  & $99.89$ { \tiny $\pm 0.06$ }  & $93.04$ { \tiny $\pm 0.14$ }  & $5.50$ { \tiny $\pm 0.39$ }  & $3.27$ { \tiny $\pm 0.13$ } 
 & 
  
  $100.00$ { \tiny $\pm 0.00$ }  & $100.00$ { \tiny $\pm 0.00$ }  & $94.62$ { \tiny $\pm 0.08$ }  & $0.40$ { \tiny $\pm 0.05$ }  & $6.77$ { \tiny $\pm 0.03$ } 
 \\\addlinespace[0.3em]

  $l_1$-Sparse & $94.29$ { \tiny $\pm 0.34$ }  & $95.63$ { \tiny $\pm 0.16$ }  & $91.55$ { \tiny $\pm 0.17$ }  & $12.03$ { \tiny $\pm 1.92$ }  & $2.26$ { \tiny $\pm 0.26$ } 
  & 
  
  $92.63$ { \tiny $\pm 0.13$ }  & $95.02$ { \tiny $\pm 0.10$ }  & $89.56$ { \tiny $\pm 0.08$ }  & $12.03$ { \tiny $\pm 0.39$ }  & $3.14$ { \tiny $\pm 0.17$ } 
 \\\addlinespace[0.3em]

  SalUn & $99.25$ { \tiny $\pm 0.12$ }  & $99.99$ { \tiny $\pm 0.01$ }  & $94.11$ { \tiny $\pm 0.13$ }  & $11.29$ { \tiny $\pm 0.56$ }  & $1.56$ { \tiny $\pm 0.20$ } 
  & 

  $95.69$ { \tiny $\pm 0.80$ }  & $97.26$ { \tiny $\pm 0.79$ }  & $91.55$ { \tiny $\pm 0.59$ }  & $11.27$ { \tiny $\pm 0.94$ }  & $3.06$ { \tiny $\pm 0.12$ } 
 \\\addlinespace[0.5em]

 \textbf{Amun} & $95.45$ { \tiny $\pm 0.19$ }  & $99.57$ { \tiny $\pm 0.00$ }  & $93.45$ { \tiny $\pm 0.22$ }  & $12.55$ { \tiny $\pm 0.08$ }  & $\bf 0.59$ { \tiny $\pm 0.09$ } 
 & 
 
  $93.50$ { \tiny $\pm 0.09$ }  & $99.71$ { \tiny $\pm 0.01$ }  & $92.39$ { \tiny $\pm 0.04$ }  & $13.53$ { \tiny $\pm 0.19$ }  & $\bf 1.37$ { \tiny $\pm 0.07$ } 
 \\\addlinespace[0.3em]

 \textbf{Amun}$_{+SalUn}$ & $94.73$ { \tiny $\pm 0.07$ }  & $99.92$ { \tiny $\pm 0.01$ }  & $93.95$ { \tiny $\pm 0.18$ }  & $14.23$ { \tiny $\pm 0.40$ }  & $\underline{0.60}$ { \tiny $\pm 0.10$ } 
 & 
 
$93.56$ { \tiny $\pm 0.07$ }  & $99.72$ { \tiny $\pm 0.02$ }  & $92.52$ { \tiny $\pm 0.20$ }  & $13.33$ { \tiny $\pm 0.10$ }  & $\underline{1.47}$ { \tiny $\pm 0.01$ } 
 \\\addlinespace[0.3em]

 

 
 
 \bottomrule
\end{tabular}
}
\end{sc}
\end{small}
\end{center}
\caption{\footnotesize {\bf Unlearning with access to $\Dr$.} Comparing different unlearning methods in unlearning $10\%$ and $50\%$ of $\D$. Avg. Gap (see \S~\ref{sec:metrics}), with MIS as the MIA score, is used for evaluation (lower is better). The lowest value is shown in bold while the second best is specified with underscore. As the results show, \amun{} outperforms all other methods by achieving lowest Avg. Gap and \amun{}$_{SalUn}$ achieves comparable results.\vspace{-1mm}}
\label{tab:mia}

\end{table*}

\begin{table*}[th!]
\begin{center}
\begin{small}
\begin{sc}
\resizebox{\textwidth}{!}{
\begin{tabular}{@{} l  c c  c  c c | c c c c c @{}}
 \toprule


& \multicolumn{5}{@{}c}{\textbf{Random Forget ($10\%$)}} & \multicolumn{5}{@{}c}{\textbf{Random Forget ($50\%$)}} \\\addlinespace[0.3em]

 \multicolumn{1}{c}{\scriptsize \textbf{}} & 
 \multicolumn{1}{c}{\scriptsize Unlearn Acc} & 
 \multicolumn{1}{c}{\scriptsize Retain Acc} & 
 \multicolumn{1}{c}{\scriptsize Test Acc} & 
 \multicolumn{1}{c}{\scriptsize MIA} & 
 \multicolumn{1}{c}{\scriptsize Avg. Gap} & 
 \multicolumn{1}{c}{\scriptsize Unlearn Acc} &
 \multicolumn{1}{c}{\scriptsize Retain Acc} & 
 \multicolumn{1}{c}{\scriptsize Test Acc} & 
 \multicolumn{1}{c}{\scriptsize  MIA} & 
 \multicolumn{1}{c}{\scriptsize Avg. Gap} 
 \\\addlinespace[0.3em]

 \cmidrule(r){2-6}
 \cmidrule(r){7-11}



 Retrain & $94.49$ {\tiny $\pm 0.20$} & $100.0 $ {\tiny $\pm 0.00$} & $94.33$ {\tiny $\pm 0.18$} & $12.53$ {\tiny $\pm 0.32$} & $0.00$  & 
 
 $92.09$ {\tiny $\pm 0.37$}  & $100.0$ {\tiny $\pm 0.00$} & $91.85$ {\tiny $\pm 0.33$} & $16.78$ {\tiny $\pm 0.37$} & $0.00$
 
 \\\addlinespace[0.3em]
  \cmidrule(r){1-11}

 RL & $100.00$ { \tiny $\pm 0.00$ }  & $100.00$ { \tiny $\pm 0.00$ }  & $94.45$ { \tiny $\pm 0.09$ }  & $3.06$ { \tiny $\pm 0.63$ }  & $3.77$ { \tiny $\pm 0.13$ } 
 & 
 
 $100.00$ { \tiny $\pm 0.00$ }  & $100.00$ { \tiny $\pm 0.00$ }  & $94.54$ { \tiny $\pm 0.11$ }  & $0.40$ { \tiny $\pm 0.03$ }  & $6.75$ { \tiny $\pm 0.02$ } 
 \\\addlinespace[0.3em]

  GA & $4.77$ { \tiny $\pm 3.20$ }  & $5.07$ { \tiny $\pm 3.54$ }  & $5.09$ { \tiny $\pm 3.38$ }  & $32.63$ { \tiny $\pm 50.85$ }  & $76.58$ { \tiny $\pm 7.73$ } 
  & 
 
  $100.00$ { \tiny $\pm 0.00$ }  & $100.00$ { \tiny $\pm 0.00$ }  & $94.57$ { \tiny $\pm 0.06$ }  & $0.35$ { \tiny $\pm 0.10$ }  & $6.77$ { \tiny $\pm 0.04$ } 
 \\\addlinespace[0.3em]

 BS & $100.00$ { \tiny $\pm 0.00$ }  & $100.00$ { \tiny $\pm 0.00$ }  & $94.48$ { \tiny $\pm 0.04$ }  & $1.11$ { \tiny $\pm 0.30$ }  & $4.27$ { \tiny $\pm 0.07$ } 
 & 
  
 $100.00$ { \tiny $\pm 0.00$ }  & $100.00$ { \tiny $\pm 0.00$ }  & $94.59$ { \tiny $\pm 0.03$ }  & $0.38$ { \tiny $\pm 0.02$ }  & $6.76$ { \tiny $\pm 0.01$ }
 \\\addlinespace[0.3em]

  SalUn  & $100.00$ { \tiny $\pm 0.00$ }  & $100.00$ { \tiny $\pm 0.00$ }  & $94.47$ { \tiny $\pm 0.10$ }  & $2.39$ { \tiny $\pm 0.64$ }  & $3.95$ { \tiny $\pm 0.14$ }
  & 

   $100.00$ { \tiny $\pm 0.00$ }  & $100.00$ { \tiny $\pm 0.00$ }  & $94.57$ { \tiny $\pm 0.12$ }  & $0.33$ { \tiny $\pm 0.04$ }  & $6.77$ { \tiny $\pm 0.03$ } 
 \\\addlinespace[0.5em]

 \textbf{Amun} & $94.28$ { \tiny $\pm 0.37$ }  & $97.47$ { \tiny $\pm 0.10$ }  & $91.67$ { \tiny $\pm 0.04$ }  & $11.61$ { \tiny $\pm 0.60$ }  & $\underline{1.61}$ { \tiny $\pm 0.09$ } 
 & 
 
 $92.77$ { \tiny $\pm 0.52$ }  & $95.66$ { \tiny $\pm 0.25$ }  & $89.43$ { \tiny $\pm 0.19$ }  & $14.13$ { \tiny $\pm 0.67$ }  & $\underline{2.52}$ { \tiny $\pm 0.16$ } 
 \\\addlinespace[0.3em]

 \textbf{Amun}$_{+Salun}$ & $94.19$ { \tiny $\pm 0.38$ }  & $97.71$ { \tiny $\pm 0.06$ }  & $91.79$ { \tiny $\pm 0.12$ }  & $11.66$ { \tiny $\pm 0.16$ }  & $\bf 1.51$ { \tiny $\pm 0.02$ } 
 & 
 
 $91.90$ { \tiny $\pm 0.63$ }  & $96.59$ { \tiny $\pm 0.31$ }  & $89.98$ { \tiny $\pm 0.44$ }  & $13.07$ { \tiny $\pm 0.66$ }  & $\bf 2.35$ { \tiny $\pm 0.15$ } 
 \\\addlinespace[0.3em]

 \bottomrule
\end{tabular}
}
\end{sc}
\end{small}
\end{center}
\caption{\footnotesize {\bf Unlearning with access to only $\Df$.}  Comparing different unlearning methods in unlearning $10\%$ and $50\%$ of $\D$. Avg. Gap (see \S~\ref{sec:metrics}) is used for evaluation (lower is better) when only $\Df$ is available during unlearning. As the results show, \amun{}$_{SalUn}$ significantly outperforms all other methods, and \amun{} achieves comparable results. \vspace{-3mm}}
\label{tab:mia_forgetonly}

\end{table*}

\section{Continuous Unlearning (cont.)}
\label{apx:adaptive}

In \S~\ref{sec:adaptive}, we showed \amun{}, whether with adaptive computation of $\Da$ or using the pre-computed ones, outperforms other unlearning methods when handling multiple unlearning requests. Another important observation on the presented results in Figure~\ref{fig:adaptive} is that {\em the effectiveness of unlearning decreases with the number of unlearning requests}. For the setting with access to $\Dr$, this decrease is due to the fact that the $\Df$ at each step has been a part of $\Dr$ at the previous steps; the model has been fine-tuned on this data in all the previous steps which has led to further improving confidence of the modes on predicting those samples. 
This result also matches the theoretical and experimental results in differential privacy literature as well~\citep{dwork2006differential,abadi2016deep}. 

This problem does not exist for the setting where there is no access to $\Dr$, but we still see a decrease in the unlearning effectiveness as we increase the number of unlearning requests. The reason behind this deterioration is that the model itself is becoming weaker. As Figure~\ref{fig:adaptive_FT_RF} shows, the accuracy on the model on both $\Dr$ and $\Dt$ gets worse as it proceeds with the unlearning request; this is because each unlearning step shows the model only $2\%$ ($1$K) of the samples and their corresponding adversarial examples for fine-tuning. So this deterioration is expected after a few unlearning requests. So when using \amun{} in this setting (no access to $\Dr$) in practice, it would be better to decrease the number of times that the unlearning request is performed, for example by performing a lazy-unlearning (waiting for a certain number of requests to accumulate) or at least using a sub-sample of $\Dr$ if that is an option.

\begin{figure*}[h!]
\centering
\includegraphics[width=0.28\textwidth]{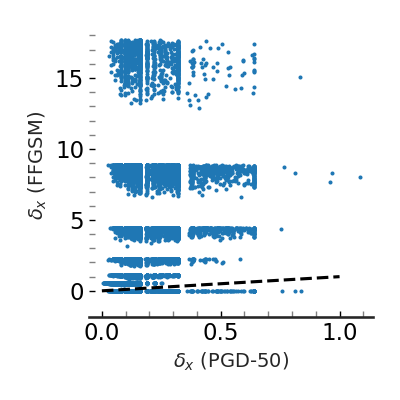}
  \caption{For each image in CIFAR-10 the $x$-axis shows the Euclidean distance of the corresponding adversarial example that is found by using PGD-50 in Algorithm~\ref{alg:advset}. $y$-axis shows this distance for the adversarial examples found by the variant of FFGSM in Algorithm~\ref{alg:advset}. The dashed line shows the $x=y$ line.}
  \label{fig:fgsm_ratio}
\end{figure*}


\end{document}